\documentclass[11pt]{article}

\usepackage[preprint]{acl}

\usepackage{times}
\usepackage{latexsym}

\usepackage[T1]{fontenc}

\usepackage[utf8]{inputenc}

\usepackage{microtype}

\usepackage{inconsolata}

\usepackage{graphicx}

\usepackage{amsmath,amsfonts,bm}

\def\eqref#1{equation~\ref{#1}}

\def\1{\bm{1}}

\DeclareMathAlphabet{\mathsfit}{\encodingdefault}{\sfdefault}{m}{sl}
\SetMathAlphabet{\mathsfit}{bold}{\encodingdefault}{\sfdefault}{bx}{n}

\usepackage{amsthm}
\usepackage{hyperref}
\usepackage{url}

\usepackage{times}
\usepackage{latexsym}
\usepackage{amsmath}
\usepackage[T1]{fontenc}

\usepackage[utf8]{inputenc}

\usepackage{microtype}
\usepackage{tcolorbox}
\usepackage{colortbl} 
\usepackage{newfloat}
\usepackage{listings}
\usepackage{inconsolata}
\usepackage{graphicx}
\usepackage{booktabs}
\usepackage{wrapfig}
\usepackage{enumitem}
\usepackage{subfigure}
\usepackage{multirow}
\usepackage{algorithm}
\usepackage{algpseudocode}
\usepackage{algorithmicx}

\definecolor{R1}{HTML}{56C596}         
\definecolor{R1-Searcher}{HTML}{4DA5D9} 
\definecolor{Search-R1}{HTML}{3C88C6}   
\definecolor{Graph-R1}{HTML}{6A0DAD}   

\newcommand{\think}[1]{\textcolor{blue}{\texttt{<think>}} #1 \textcolor{blue}{\texttt{</think>}}}
\newcommand{\search}[1]{\textcolor{orange}{\texttt{<query>}} #1 \textcolor{orange}{\texttt{</query>}}}
\newcommand{\info}[1]{\textcolor{teal}{\texttt{<knowledge>}} #1 \textcolor{teal}{\texttt{</knowledge>}}}
\newcommand{\answer}[1]{\textcolor{purple}{\texttt{<answer>}} #1 \textcolor{purple}{\texttt{</answer>}}}

\title{Multi-hop Reasoning via Early Knowledge Alignment}

\author{{\bf Yuxin Wang\thanks{Equal contribution.}\textsuperscript{\rm 1,2}, Shicheng Fang\footnotemark[1]\textsuperscript{\rm 1,3}\ , Bo Wang\textsuperscript{\rm 1}, Qi Luo\textsuperscript{\rm 1},} \\
{ {\bf Xuanjing Huang\textsuperscript{\rm 1,2}, Yining Zheng\textsuperscript{\rm 1}, Xipeng Qiu\textsuperscript{\rm 1,3}}} \\
         \textsuperscript{1}Computer Science, Fudan University\ \ \\ \textsuperscript{2}Institute
of Modern Languages and Linguistics, Fudan University \\
         \textsuperscript{3}Shanghai SII\ \  \\ 
         \texttt{\{wangyuxin21,25113050022,22110240036,qluo22\}@m.fudan.edu.cn} \\
         \texttt{\{ynzheng19,xjhuang,xpqiu\}@fudan.edu.cn}}

\begin{document}
\maketitle
\begin{abstract}

Retrieval-Augmented Generation (RAG) has emerged as a powerful paradigm for Large Language Models (LLMs) to address knowledge-intensive queries requiring domain-specific or up-to-date information. To handle complex multi-hop questions that are challenging for single-step retrieval, iterative RAG approaches incorporating reinforcement learning have been proposed. However, existing iterative RAG systems typically plan to decompose questions without leveraging information about the available retrieval corpus, leading to inefficient retrieval and reasoning chains that cascade into suboptimal performance. In this paper, we introduce Early Knowledge Alignment (EKA), a simple but effective module that aligns LLMs with retrieval set before planning in iterative RAG systems with contextually relevant retrieved knowledge. Extensive experiments on six standard RAG datasets demonstrate that by establishing a stronger reasoning foundation, EKA significantly improves retrieval precision, reduces cascading errors, and enhances both performance and efficiency. Our analysis from an entropy perspective demonstrate that incorporating early knowledge reduces unnecessary exploration during the reasoning process, enabling the model to focus more effectively on relevant information subsets. Moreover, EKA proves effective as a versatile, training-free inference strategy that scales seamlessly to large models. Generalization tests across diverse datasets and retrieval corpora confirm the robustness of our approach. Overall, EKA advances the state-of-the-art in iterative RAG systems while illuminating the critical interplay between structured reasoning and efficient exploration in reinforcement learning-augmented frameworks. The code is released at \href{https://github.com/yxzwang/EarlyKnowledgeAlignment}{Github}.

\end{abstract}

\section{Introduction}

\begin{figure}[tbp]
    \centering
    \includegraphics[width=0.9\linewidth]{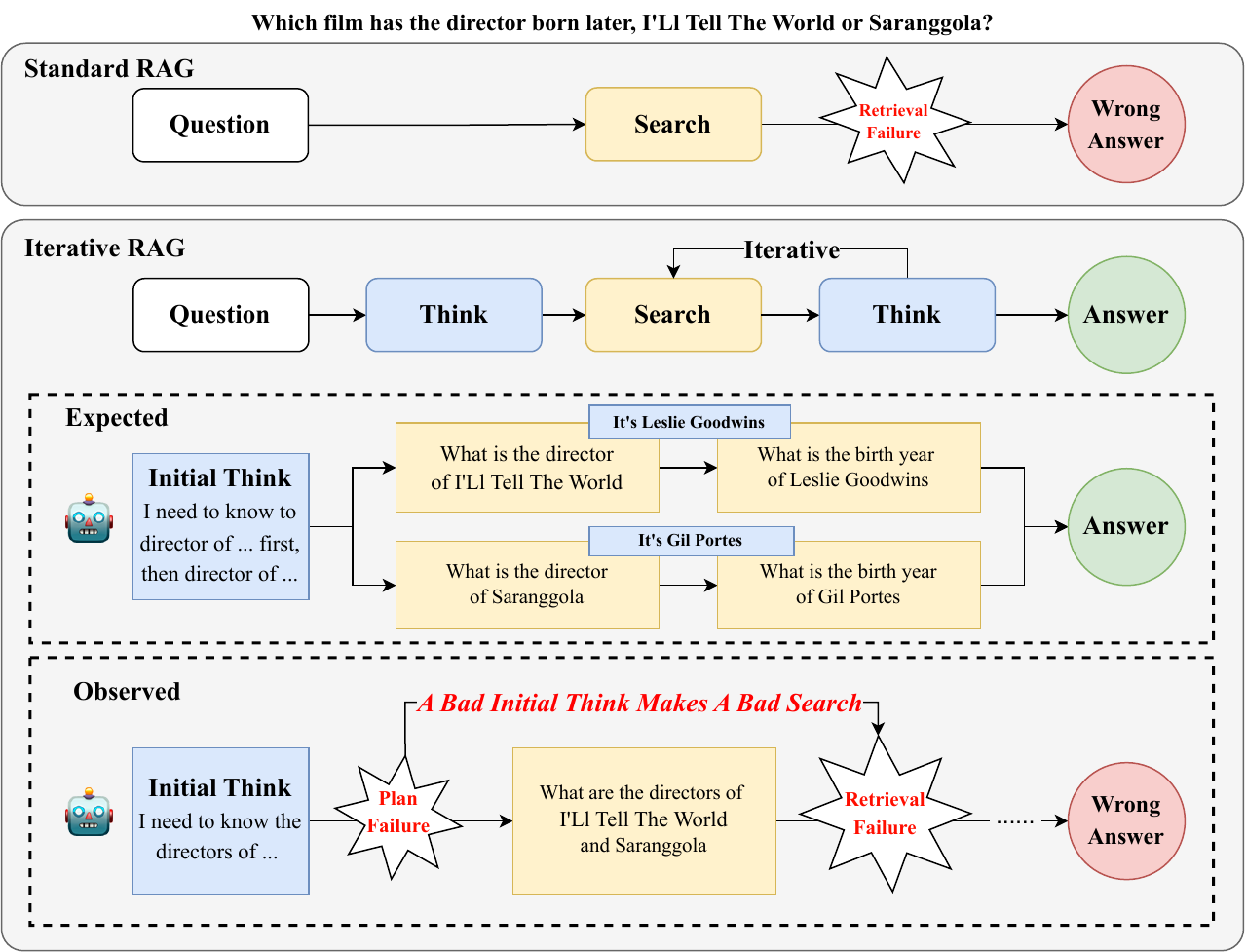}
    \vspace{-3mm}
    \caption{Standard RAG and Iterative RAG pipeline. While standard RAG suffers from the impossibility of multi-hop retrieval in one step, iterative RAG also suffers from plan failure in the initial think, which is caused by lack of information of the retrieval set.}
    \label{fig:EKA_introduction}
    \vspace{-1.5mm}
\end{figure}

Large Language Models (LLMs) have demonstrated remarkable capabilities in natural language understanding and generation, yet they face fundamental limitations when dealing with knowledge-intensive tasks that require access to up-to-date or domain-specific information. Retrieval-Augmented Generation (RAG) has emerged as a promising paradigm to address these limitations by dynamically incorporating external knowledge from retrieval corpora into the generation process \citep{karpukhin2020dense,RAG}. Standard RAG systems perform a single retrieval step followed by generation, but the intrinsic difficulty of retrieving multi-hop information in one step causes a lot of failure. Recent advances have shown that iterative approaches where models can perform multiple rounds of retrieval and reasoning—significantly improve performance on complex multi-hop reasoning tasks\citep{Search-R1,DeepRAG,Graph-R1,R1-Searcher}. However, although assumed well, these iterative systems can still suffer from retrieval failure, resulting from the plan failure which leads to the suboptimal reasoning chains, particularly when the initial reasoning step lacks sufficient contextual grounding. These scenarios are illustrated in Figure \ref{fig:EKA_introduction} with a real example from the dataset.

Iterative RAG systems\citep{Search-R1,R1-Searcher} are often optimized by Reinforcement Learning (RL)\citep{PPO,GRPO}, offering a principled approach to learn effective retrieval and reasoning strategies. RL-based RAG frameworks treat the retrieval and generation process as a sequential decision-making problem, where agents learn to search for information and generate responses to maximize cumulative rewards based on answer accuracy and efficiency metrics. The success of RL training heavily depends on the quality of the exploitation and the exploration efficiency during the learning process. Recent studies on entropy\citep{wang2025beyond,cui2025entropy} show that entropy measurement is a good signal for this exploitation and exploration balance, which is important because the exploitation of retrieved information and exploration in the retrieval set control the whole reasoning process. Poor initial reasoning steps in exploration can lead to compounding errors throughout the iterative process. 

From both the perspective of an iterative RAG system and the RL training dynamics, the quality of initial planning plays a crucial role in the effectiveness of generating right answers. When models begin their reasoning process without adequate contextual knowledge, they often generate misguided hypotheses or pursue irrelevant reasoning paths relying on themselves, which is far from the information the environment can give, leading to a cascade of poor retrieval decisions and incorrect conclusions. This problem is particularly pronounced in the early stages of RL training, where random or poorly informed initial actions can significantly hinder the learning process. By enhancing the initial planning step with early knowledge, we hypothesize that models can establish more accurate reasoning foundations, leading to better exploration strategies with less entropy and more efficient learning dynamics. This \textbf{Early Knowledge Alignment (EKA)} not only improves the immediate reasoning quality but also provides clearer learning signals for the RL algorithm, enabling faster roads to the right answer.

Our contribution is as follows:
\vspace{-1.5mm}
\begin{itemize}
    \item \textbf{Early Knowledge Alignment (EKA).} We propose a novel approach that augments the initial thinking step in iterative RAG systems with early knowledge, providing models with better grounding before entering the RL-optimized iterative retrieval and generation process. This framework significantly improves the quality of reasoning foundations and reduces the likelihood of cascading errors in subsequent iterations.
    \vspace{-1.5mm}
    \item \textbf{Analysis from an Entropy Perspective.} We analyze the training dynamics of Group Relative Policy Optimization (GRPO)\citep{GRPO} in iterative RAG from an entropy perspective and show that with lower entropy in the training phase, instead of insufficient exploration, our approach leads to more efficient exploration strategies focusing on the retrieval set, faster roads to the answer during RL training compared to traditional approaches that start with uninformed, model initialized thinking.
    \vspace{-1.5mm}
    \item \textbf{Comprehensive Experimental Validation.} We conduct extensive experiments on standard RAG datasets, showing consistent improvements in both answer accuracy and retrieval recall. Besides, generalization experiments show no degrading of generalization with our method.
    
\end{itemize}

\vspace{-5mm}
\section{Related Works}
\vspace{-1.5mm}
\subsection{Retrieval-Augmented Generation}

The concept of augmenting language models with external knowledge retrieval has gained significant traction in recent years. Early work by \citep{karpukhin2020dense} introduced Dense Passage Retrieval (DPR), which demonstrated the effectiveness of dense vector representations for retrieval in open-domain question answering. \citep{RAG} proposed Retrieval-Augmented Generation and a lot of works\citep{alce, udr} has emerged. To apply better retrieval, LightRAG\citep{LightRAG} employs a dual-level retrieval system for better generation. Structure-based retrieval methods like GraphRAG\citep{GraphRAG}, PathRAG\citep{PathRAG}, HippoRAG2\citep{HippoRAG2}, HyperGraphRAG\citep{HyperGraphRAG} have been proposed to utilize fine-grained retrieval like entities or links and generate better responses. Traditional single-step RAG systems often fall short when dealing with complex reasoning tasks that require multiple pieces of evidence. This limitation has motivated research into iterative RAG systems.

\subsection{Iterative and Multi-Hop RAG Approaches}

Chain-of-Thought (CoT) prompting \citep{wei2022chain} encourages models to generate intermediate reasoning steps, effectively simulating an iterative thinking process. IRCoT \citep{trivedi2022interleaving} demonstrated that interleaving retrieval and generation steps can significantly improve performance on multi-hop reasoning tasks. ITER-RETGEN \citep{shao2023enhancing} proposed a framework where models can decide when to retrieve additional information based on their confidence levels. WebGPT \citep{nakano2021webgpt} showed that models can be trained to browse the web iteratively to gather information for answering questions. ReAct \citep{yao2023react} combined reasoning and acting in language models, enabling them to perform dynamic retrieval based on their reasoning traces. More recent work by \citep{asai2024self} introduced Self-RAG, which uses reflection tokens to control retrieval timing and assess the quality of retrieved passages,while Self-ask, proposed by \citep{press2022measuring}, implements an autonomous question formulation mechanism during the reasoning process. FLARE \citep{jiang2023active} incorporates adaptive retrieval when LLMs generate low-confidence tokens.

\vspace{-3mm}
\subsection{Reinforcement Learning for RAG Optimization}

The application of reinforcement learning to optimize RAG systems has emerged as a promising research direction. Several approaches, such as R1-Searcher\citep{R1-Searcher}, R3-RAG\citep{r3-rag}, and DeepRAG\citep{DeepRAG}, employ a two-stage training process. They first use manually curated data to perform Supervised Fine-Tuning (SFT) on the LLM, and subsequently apply reinforcement learning to further align the model with the available knowledge boundaries. Similarly, s3 \citep{jiang2025s3} proposes a modular framework that employs RL to optimize a search agent while keeping the generator frozen, focusing on input context optimization rather than joint reasoning. A critical problem is that some multi-hop questions have more than one good reasoning paths, which requires high quality for sft data. Search-R1\citep{Search-R1}, DeepResearcher\citep{zheng-etal-2025-deepresearcher} and Graph-R1\citep{Graph-R1} directly applies reinforcement learning on LLMs. Consequently, these approaches rely more heavily on the LLM's innate reasoning capabilities to solve the questions without a preceding SFT stage. This may introduce redundant paths when LLM does not align with the retrieval set. Our method applies Early Knowledge Alignment to alleviate this problem.

\section{Preliminaries}
\vspace{-1.5mm}
\subsection{PPO}
\vspace{-1.5mm}
Proximal Policy Optimization (PPO) \citep{PPO} is an actor-critic reinforcement learning algorithm that has become the predominant method for RL fine-tuning of large language models \citep{ouyang2022training}. For language model fine-tuning, PPO maximizes the following objective:

\begin{align}
    &\mathcal{J}_{PPO}(\theta) = \mathbb{E}_{[q \sim P(Q), o \sim \pi_{\theta_{old}}(O|q)]} \\ &\left[ \frac{1}{|o|} \sum_{t=1}^{|o|} \min \left( r_t(\theta) A_{t}, \text{clip}(r_t(\theta), 1 - \epsilon, 1 + \epsilon) A_{t} \right) \right],
\end{align}
where $r_t(\theta) = \frac{\pi_\theta(o_{t} | q, o_{<t})}{\pi_{\theta_{old}}(o_{t} | q, o_{<t})}$ is the probability ratio between the current policy $\pi_{\theta}$ and the old policy $\pi_{\theta_{old}}$. Here, $q$ and $o$ represent questions sampled from the dataset $P(Q)$ and corresponding outputs generated by the old policy, respectively. The clipping parameter $\epsilon$ constrains the policy ratio to the interval $[1-\epsilon, 1+\epsilon]$, preventing destabilizing updates. $A_t$ denotes the advantage function, typically computed using Generalized Advantage Estimation (GAE)\citep{gae} based on rewards and a learned value function $V_{\psi}$.

\vspace{-1.5mm}
\subsection{GRPO}
\vspace{-1.5mm}

\citep{GRPO} propose Group Relative Policy Optimization (GRPO), illustrated in Figure \ref{fig:GRPOwithInitialKnowledge}. GRPO eliminates the need for value function approximation by using the average reward of multiple sampled outputs as a baseline. For each question $q$, GRPO samples a group of $G$ outputs $\{o_1, o_2, \ldots, o_G\}$ from the old policy $\pi_{\theta_{old}}$ and optimizes the following objective:
\begin{equation}
\begin{aligned}
\mathcal{J}_{\text{GRPO}}(\theta)
= \mathbb{E}_{q \sim P(Q),\, \{o_i\}_{i=1}^G \sim \pi_{\theta_{\text{old}}}(O|q)} \\
\Bigg[
\frac{1}{G} \sum_{i=1}^G \frac{1}{|o_i|} \sum_{t=1}^{|o_i|}
\Big(
\min\big(r_t(\theta)\hat{A}_{i,t},\\
\, \operatorname{clip}(r_t(\theta),\, 1-\varepsilon,\, 1+\varepsilon)\hat{A}_{i,t}\big) \\
- \beta\, \mathbb{D}_{\text{KL}}(\pi_{\theta} \,||\, \pi_{\text{ref}})
\Big)
\Bigg].
\end{aligned}
\label{eq:GRPO-obj}
\end{equation}

where $r_{i,t}(\theta) = \frac{\pi_\theta(o_{i,t} | q, o_{i,<t})}{\pi_{\theta_{old}}(o_{i,t} | q, o_{i,<t})}$ is the probability ratio, and $\hat{A}_{i,t}$ represents the advantage computed using relative rewards within each group: 
\begin{equation}
    \hat{A}_{i, t} = \widetilde{r}_i = \frac{r_i- {\rm mean}(\mathbf{r})}{{\rm std}(\mathbf{r})}
\end{equation}
 where $\mathbf{r}=\{r_1, r_2, \cdots, r_G\}$ is the rewards tensor of $G$ samples in the group correspondingly. The group-relative advantage computation aligns naturally with how reward models are trained—on comparative datasets where outputs for the same question are ranked against each other.

\vspace{-1.5mm}
\section{Method}
\vspace{-1.5mm}

We propose \textbf{Early Knowledge Alignment (EKA)}, a simple but effective module that enhances iterative RAG systems by incorporating early knowledge before the initial planning. Our method addresses the fundamental limitation of normal planning, in all existing iterative RAG systems where models begin reasoning without sufficient contextual grounding, often leading to suboptimal retrieval strategies and redundant exploration during reinforcement learning.

\begin{figure}[htbp]
    \centering
    \includegraphics[width=0.9\linewidth]{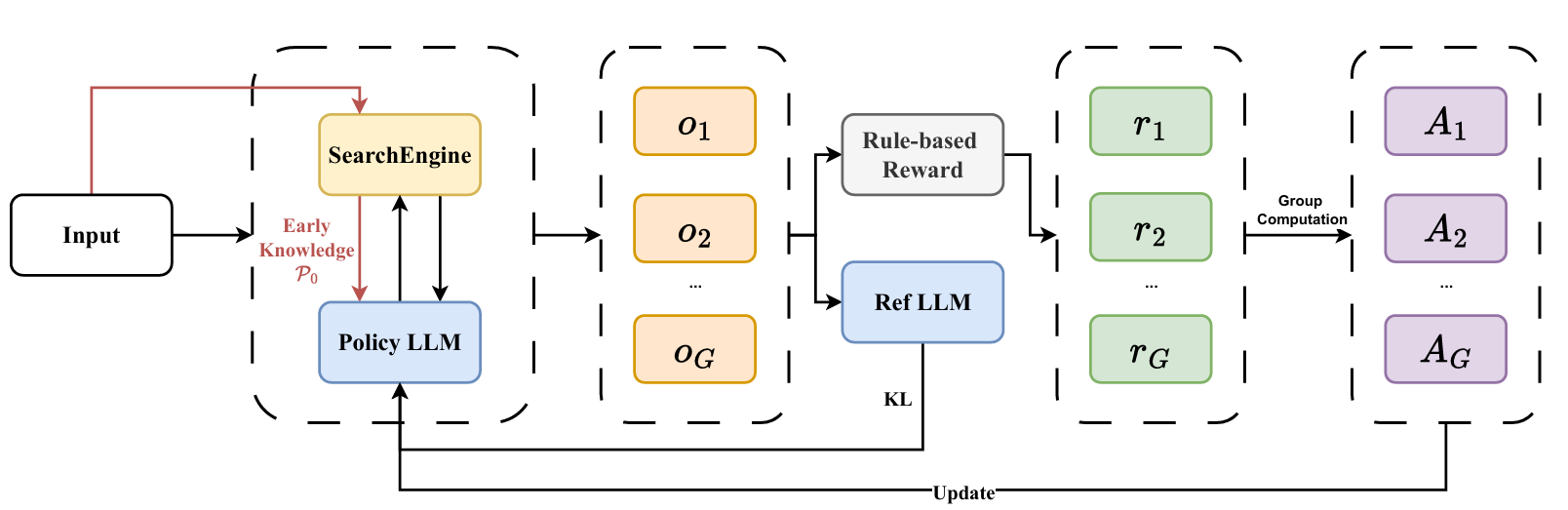}
    \caption{GRPO training with EKA.}
    \label{fig:GRPOwithInitialKnowledge}
\end{figure}

Figure~\ref{fig:GRPOwithInitialKnowledge} illustrates the GRPO training pipeline of EKA. The policy LLM receives Early Knowledge $\mathcal{P}_0$ from the SearchEngine before its first thinking step. Subsequently, the model proceeds with the standard rollout and update phases as in conventional GRPO training. Algorithm is referred to Appendix \ref{sec:appendix-algo}.
\vspace{-1.5mm}
\subsection{Early Knowledge Alignment}

Given an input question $q$, our EKA approach first performs an initial retrieval step to gather relevant knowledge before generating the initial thinking step. Specifically, we retrieve the top-$k$ most relevant passages from the knowledge corpus $\mathcal{D}$ using a retriever:

\begin{equation}
\mathcal{P}_0 = \text{Retrieve}(q, \mathcal{D}, k),
\end{equation}

where $\mathcal{P}_0 = \{p_1, p_2, \ldots, p_k\}$ represents the initially retrieved passages.

\subsection{Iterative Thinking and Searching}

Following the initial search, our method proceeds with iterative thinking and searching, now grounded by early knowledge, until a final answer is generated. The action pipeline is set as $[a_0, a_1, a_2,... a_t]$ where $a_0$ is \textbf{Search} and at each subsequent step $i > 0$, action $a_i$ is \textbf{Search} or \textbf{Answer} $\text{if } a_{i-1} = \textbf{Think}$ and \textbf{Think}  $\text{if } a_{i-1} != \textbf{Think}$. Each action is defined as:
\begin{itemize}
\item \textbf{Think}: Generate reasoning steps based on current knowledge.
\item \textbf{Search}: Query the knowledge base for additional information.
\item \textbf{Answer}: Provide the final answer when sufficient information is gathered.
\end{itemize}
To guide the model in producing this sequence of actions, we employ the prompt detailed in Table  \ref{tab:promptofEKA}, which instructs it to generate structured outputs.

\begin{table}[htbp]
\centering
\small
\caption{Template for the updated prompt. Note that early knowledge is provided within \textcolor{teal}{\texttt{\textbf{<knowledge>}}}...\textcolor{teal}{\texttt{\textbf{</knowledge>}}} at the beginning, and additional retrieved knowledge is placed within the same tags after \textcolor{orange}{\texttt{\textbf{</query>}}}.}\label{tab:promptofEKA}
\resizebox{0.5\textwidth}{!}{\begin{tabular}{p{13cm}}
\toprule
Answer the given question. You can query from knowledge base provided to you to answer the question. You can query knowledge as many times as you want. The initial knowledge you need for the first think is between \textcolor{teal}{\texttt{\textbf{<knowledge>}}}...\textcolor{teal}{\texttt{\textbf{</knowledge>}}}.
You must first conduct reasoning inside \textcolor{blue}{\texttt{\textbf{<think>}}}...\textcolor{blue}{\texttt{\textbf{</think>}}} relied on the initial knowledge. If you need to query knowledge, you can set a query statement between 
\textcolor{orange}{\texttt{\textbf{<query>}}}...\textcolor{orange}{\texttt{\textbf{</query>}}} 
to query from knowledge base after \textcolor{blue}{\texttt{\textbf{<think>}}}...\textcolor{blue}{\texttt{\textbf{</think>}}}.  
When you have the final answer, you can output the answer inside 
\textcolor{purple}{\texttt{\textbf{<answer>}}}...\textcolor{purple}{\texttt{\textbf{</answer>}}}. Question: \textcolor{red}{question}. \textcolor{teal}{\texttt{\textbf{<knowledge>}}}\textcolor{teal}{Knowledge}\textcolor{teal}{\texttt{\textbf{</knowledge>}}}. Assistant: \\
\bottomrule
\end{tabular}}

\vspace{-3mm}  
\end{table}

\begin{table*}[tbp]
\centering
\caption{\label{T2} Main results in Graph-R1 setting with best in \textbf{bold}. \raisebox{-0.5mm}{\includegraphics[width=0.02\textwidth]{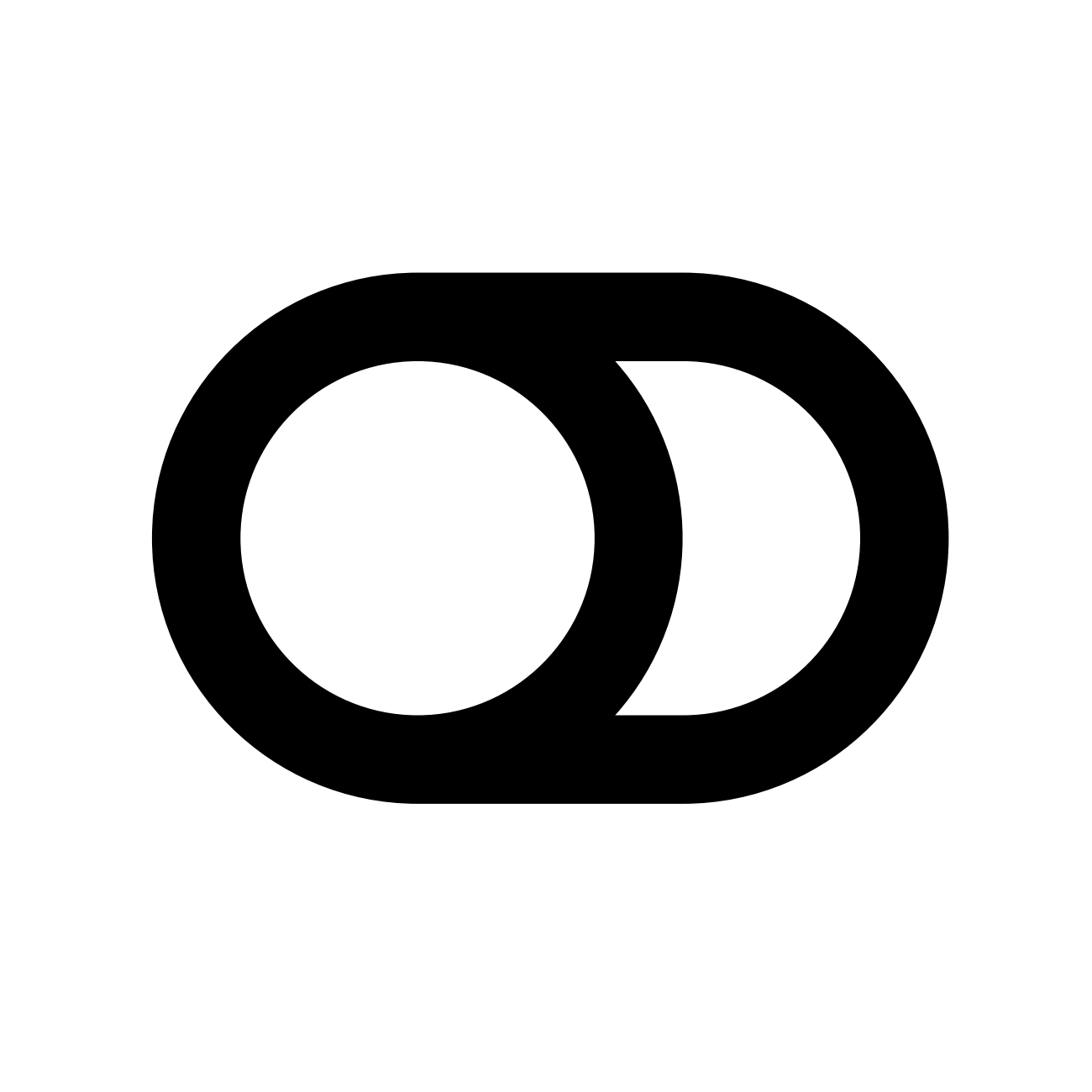}} means prompt engineering, \raisebox{-0.5mm}{\includegraphics[width=0.02\textwidth]{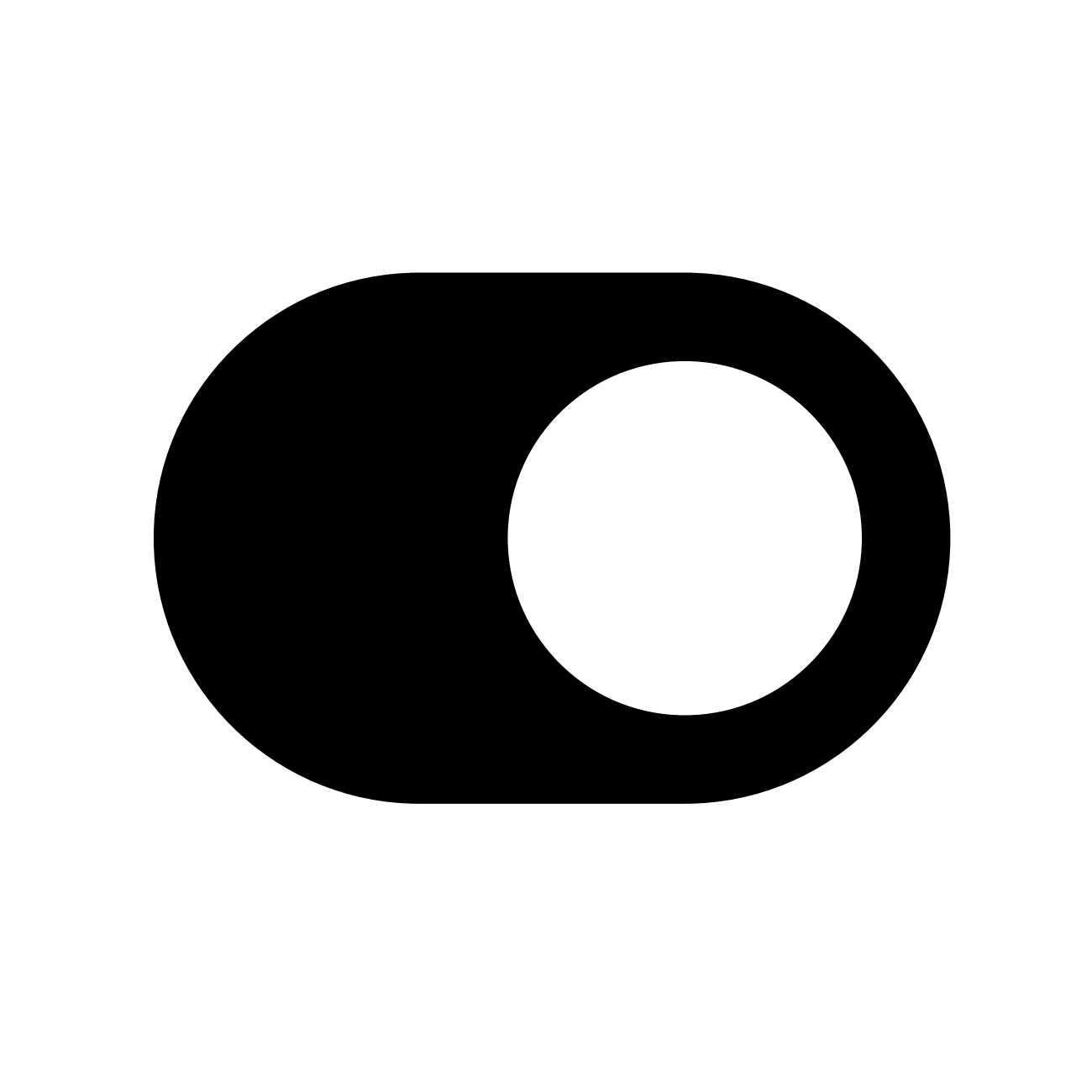}} means training, \raisebox{-0.5mm}{\includegraphics[width=0.02\textwidth]{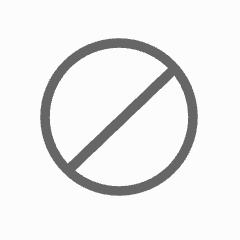}} means no knowledge interaction, \raisebox{-0.5mm}{\includegraphics[width=0.02\textwidth]{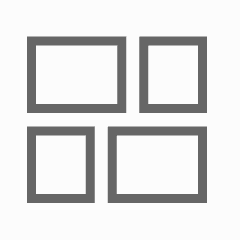}} means chunk-based knowledge, and \raisebox{-0.5mm}{\includegraphics[width=0.02\textwidth]{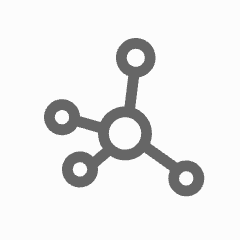}} means graph-based knowledge.}\label{tab:graphr1}
\fontsize{7pt}{7.5pt}\selectfont
\setlength{\tabcolsep}{0.8mm}{
\resizebox{\textwidth}{!}{\begin{tabular}{l cc  cc cc cc cc cc |ccc}
\toprule
\multirow{2.5}{*}{\textbf{Method}} & \multicolumn{2}{c}{\textbf{2Wiki.}} & \multicolumn{2}{c}{\textbf{HotpotQA}} & \multicolumn{2}{c}{\textbf{Musique}} & \multicolumn{2}{c}{\textbf{NQ}} & \multicolumn{2}{c}{\textbf{PopQA}} & \multicolumn{2}{c}{\textbf{TriviaQA}} & \multicolumn{3}{c}{\textbf{Avg.}} \\
\cmidrule(lr){2-3} \cmidrule(lr){4-5} \cmidrule(lr){6-7} \cmidrule(lr){8-9} \cmidrule(lr){10-11} \cmidrule(lr){12-13} \cmidrule(lr){14-16}
 & EM & F1 & EM & F1 & EM & F1 & EM & F1 & EM & F1 & EM & F1 & EM & F1 & R-S \\
\midrule
\multicolumn{16}{c}{\textbf{\textit{GPT-4o-mini}}} \\
\raisebox{-0.22\height}{\includegraphics[width=0.02\textwidth]{close.png}}\raisebox{-0.22\height}{\includegraphics[width=0.02\textwidth]{none.png}} NaiveGeneration & 4.69& 17.03& 18.75& 31.79& 3.13& 11.45& 2.34& 21.59& 10.36& 25.95& 28.91& 47.73
& 11.36 & 25.92 & - \\
\raisebox{-0.22\height}{\includegraphics[width=0.02\textwidth]{close.png}}\raisebox{-0.22\height}{\includegraphics[width=0.02\textwidth]{chunk.png}} StandardRAG & 7.03& 22.31& 35.16& 46.70& 9.38& 17.31& 7.03& 26.85& 18.75& 30.58& 31.25& 48.55
& 18.10 & 32.05 & 52.68 \\
\raisebox{-0.22\height}{\includegraphics[width=0.02\textwidth]{close.png}}\raisebox{-0.22\height}{\includegraphics[width=0.02\textwidth]{graph.png}} GraphRAG & 3.91& 16.02& 19.53& 31.67& 7.03& 15.14& 3.91& 20.31& 8.59& 20.92& 32.03& 45.13
& 12.50 & 24.87 & 32.48  \\
\raisebox{-0.22\height}{\includegraphics[width=0.02\textwidth]{close.png}}\raisebox{-0.22\height}{\includegraphics[width=0.02\textwidth]{graph.png}} LightRAG & 3.13& 16.59& 18.75& 30.70& 3.91& 14.39& 2.34& 19.09& 5.47& 24.47& 25.00& 40.18
& 9.77 & 24.24 & 47.42\\
\raisebox{-0.22\height}{\includegraphics[width=0.02\textwidth]{close.png}}\raisebox{-0.22\height}{\includegraphics[width=0.02\textwidth]{graph.png}} PathRAG & 3.91& 12.42& 10.94& 23.12& 3.13& 11.49& 2.34& 20.01& 2.34& 15.65& 19.53& 37.44
& 7.03 & 20.02 & 46.71  \\
\raisebox{-0.22\height}{\includegraphics[width=0.02\textwidth]{close.png}}\raisebox{-0.22\height}{\includegraphics[width=0.02\textwidth]{graph.png}} HippoRAG2 & 7.03& 16.27 & 19.53 & 31.78 & 6.25& 12.37& 7.81& 24.56& 9.38& 21.10& 32.81& 48.86
& 13.80 & 25.82 & 36.41 \\
\raisebox{-0.22\height}{\includegraphics[width=0.02\textwidth]{close.png}}\raisebox{-0.22\height}{\includegraphics[width=0.02\textwidth]{graph.png}} HyperGraphRAG & 4.69 & 21.14 & 21.88 & 37.46 & 6.25 & 20.40 & 3.91 & 22.95 & 13.28 & 29.48 & 28.91 & 44.95 & 13.15 & 29.40 & 61.82 \\
\midrule
\multicolumn{16}{c}{\textbf{\textit{Qwen2.5-7B-Instruct}}} \\
\raisebox{-0.22\height}{\includegraphics[width=0.02\textwidth]{close.png}}\raisebox{-0.22\height}{\includegraphics[width=0.02\textwidth]{none.png}} NaiveGeneration & 3.12& 12.25& 6.25& 18.58& 0.00& 4.06& 1.56& 13.00& 0.78& 12.82& 7.03& 24.51
& 3.12 & 14.20 & -  \\
\raisebox{-0.22\height}{\includegraphics[width=0.02\textwidth]{close.png}}\raisebox{-0.22\height}{\includegraphics[width=0.02\textwidth]{chunk.png}} StandardRAG & 7.81& 12.75& 10.16& 21.10& 0.78& 4.53& 1.56& 15.97& 3.12& 13.10& 8.59& 24.90
& 5.34 & 15.39 & 52.67  \\
\raisebox{-0.22\height}{\includegraphics[width=0.02\textwidth]{open.png}}\raisebox{-0.22\height}{\includegraphics[width=0.02\textwidth]{none.png}} SFT & 11.72& 20.28& 19.53& 27.59& 5.47& 10.02& 5.12& 19.02& 20.31& 27.93& 31.25& 39.21
& 15.57 & 24.01 & -  \\
 \raisebox{-0.22\height}{\includegraphics[width=0.02\textwidth]{open.png}}\raisebox{-0.22\height}{\includegraphics[width=0.02\textwidth]{none.png}} R1 & 25.00& 30.99& 31.25& 37.05& 7.03& 14.53& 16.41& 28.45& 26.56& 30.35& 49.22& 57.33
& 25.91 & 33.12 & -  \\
\raisebox{-0.22\height}{\includegraphics[width=0.02\textwidth]{open.png}}\raisebox{-0.22\height}{\includegraphics[width=0.02\textwidth]{chunk.png}} R1-Searcher & 27.34& 33.96& 39.84& 46.36& 10.16& 16.63& 32.03& 44.93& 41.41& 47.12& 56.25& 64.76
& 34.51 & 42.29 & 51.26  \\
\rowcolor{R1!15}\raisebox{-0.22\height}{\includegraphics[width=0.02\textwidth]{open.png}}\raisebox{-0.22\height}{\includegraphics[width=0.02\textwidth]{chunk.png}} Search-R1 & 35.15& 38.21& 43.77& 51.26& 17.18& 21.45& \textbf{38.34}& 43.79& 43.75& 47.03& 51.56& 61.03& 38.29 & 43.80 & 53.06 
\\
\rowcolor{R1!15} \raisebox{-0.22\height}{\includegraphics[width=0.02\textwidth]{open.png}}\raisebox{-0.22\height}{\includegraphics[width=0.02\textwidth]{chunk.png}} \quad\quad\quad\quad + EKA & 56.25&60.75& 54.68& 60.44& 32.81& 41.54& 34.37& 48.97& 46.87& 51.17& 62.50& 69.79& 47.91 & 55.44 & \textbf{65.02}
\\
\rowcolor{R1!15} \raisebox{-0.22\height}{\includegraphics[width=0.02\textwidth]{open.png}}\raisebox{-0.22\height}{\includegraphics[width=0.02\textwidth]{graph.png}} \quad\quad\quad\quad $\Delta$ & \emph{+21.10}& \emph{+22.54}& \emph{+10.91}& \emph{+9.18}& \emph{+15.63}& \emph{+20.09}& \emph{-3.97}& \emph{+5.18}& \emph{+3.12}& \emph{+4.14}& \emph{+10.94}& \emph{+8.76}& \emph{+9.62} & \emph{+11.64} & \emph{+11.96} 
\\
\rowcolor{Search-R1!15} \raisebox{-0.22\height}{\includegraphics[width=0.02\textwidth]{open.png}}\raisebox{-0.22\height}{\includegraphics[width=0.02\textwidth]{chunk.png}} Search-R1-PPO & 39.84& 42.38& 47.66& 56.28& 21.09& 32.91& 18.75& 32.27& 39.08& 44.26& 60.15& 69.29& 37.76 & 46.23 & 49.31 
\\
\rowcolor{Search-R1!15} \raisebox{-0.22\height}{\includegraphics[width=0.02\textwidth]{open.png}}\raisebox{-0.22\height}{\includegraphics[width=0.02\textwidth]{chunk.png}} \quad\quad\quad\quad + EKA& 57.03&61.47& 52.34& 57.83& 30.47& 35.32& 33.59& 46.84& \textbf{49.22}& 52.34& 61.71& 69.62& 47.39 & 53.90 & \textbf{65.02} 
\\
\rowcolor{Search-R1!15} \raisebox{-0.22\height}{\includegraphics[width=0.02\textwidth]{open.png}}\raisebox{-0.22\height}{\includegraphics[width=0.02\textwidth]{graph.png}} \quad\quad\quad\quad $\Delta$ & \emph{+17.19} & \emph{+19.09}& \emph{+4.68}& \emph{+1.55}& \emph{+9.38}& \emph{+2.41}& \emph{+14.84}& \emph{+14.57}& \emph{+10.14}& \emph{+8.08}& \emph{+1.56}& \emph{+0.33}& \emph{+9.63} & \emph{+7.67} & \emph{+15.71} 
\\
\rowcolor{Graph-R1!15} \raisebox{-0.22\height}{\includegraphics[width=0.02\textwidth]{open.png}}\raisebox{-0.22\height}{\includegraphics[width=0.02\textwidth]{graph.png}} Graph-R1  & 55.47& 65.04& 57.03& 62.69& 36.72& 46.17& 33.59& 49.87& 45.31& 51.22& 63.28& 71.93& 48.57& 57.82& 60.40
\\

\rowcolor{Graph-R1!15} \raisebox{-0.22\height}{\includegraphics[width=0.02\textwidth]{open.png}}\raisebox{-0.22\height}{\includegraphics[width=0.02\textwidth]{graph.png}} \quad\quad\quad\quad + EKA & \textbf{60.94}& \textbf{68.26}& \textbf{59.38}& \textbf{66.14}& \textbf{40.63}& \textbf{51.63}& 38.28& \textbf{51.99}& 49.21& \textbf{53.49}& \textbf{64.06}& \textbf{72.37}& \textbf{52.08} & \textbf{60.65} & 64.90 
\\
\rowcolor{Graph-R1!15} \raisebox{-0.22\height}{\includegraphics[width=0.02\textwidth]{open.png}}\raisebox{-0.22\height}{\includegraphics[width=0.02\textwidth]{graph.png}} \quad\quad\quad\quad $\Delta$ & \emph{+5.47}& \emph{+3.22}& \emph{+2.35}& \emph{+3.45}& \emph{+3.91}& \emph{+5.46}& \emph{+4.69}& \emph{+2.12}& \emph{+3.90}& \emph{+2.27}& \emph{+0.78}& \emph{+0.44}& \emph{+3.51} & \emph{+2.83} & \emph{+4.50} 
\\
\midrule
\multicolumn{16}{c}{\textbf{\textit{Qwen2.5-14B-Instruct}}} \\
\rowcolor{Graph-R1!15} \raisebox{-0.22\height}{\includegraphics[width=0.02\textwidth]{open.png}}\raisebox{-0.22\height}{\includegraphics[width=0.02\textwidth]{graph.png}} Graph-R1 & 67.97& 75.46& 67.19& 72.52& 43.75& 57.54& 39.84& 53.81& 49.22& 53.33& 68.75& 76.43& 56.12& 64.85& 60.65
\\

\rowcolor{Graph-R1!15} \raisebox{-0.22\height}{\includegraphics[width=0.02\textwidth]{open.png}}\raisebox{-0.22\height}{\includegraphics[width=0.02\textwidth]{graph.png}} \quad\quad\quad\quad + EKA & \textbf{70.31}& \textbf{77.12}& \textbf{68.75}& \textbf{74.47}& \textbf{45.31}& \textbf{57.88}& \textbf{40.63}& \textbf{56.02}& \textbf{50.00}& \textbf{54.06}& \textbf{71.09}& \textbf{77.84}& \textbf{57.68} & \textbf{66.23} & \textbf{65.13}
\\
\rowcolor{Graph-R1!15} \raisebox{-0.22\height}{\includegraphics[width=0.02\textwidth]{open.png}}\raisebox{-0.22\height}{\includegraphics[width=0.02\textwidth]{graph.png}} \quad\quad\quad\quad $\Delta$ & \emph{+2.34} & \emph{+1.66} & \emph{+1.56} & \emph{+1.95} & \emph{+1.56} & \emph{+0.34} & \emph{+0.79} & \emph{+2.21} & \emph{+0.78} & \emph{+0.73} & \emph{+2.34} & \emph{+1.41} & \emph{+1.56} & \emph{+1.38} & \emph{+4.48} \\
\bottomrule
\end{tabular}}%
}
\end{table*}

\subsection{Theoretical Analysis}
In this section we propose the following proposition:

\textbf{Proposition 1.} Early Knowledge Alignment is better than traditional thinking in iterative RAG from an entropy perspective.
\begin{proof}
    The formal proof is provided in Appendix \ref{appendix:theoreticalproof}, and the empirical results regarding entropy are presented in Section \ref{ablation:entropy}.
\end{proof}

\vspace{-1.5mm}
\section{Experiments}
\vspace{-1.5mm}
\begin{table}[htbp]
    \centering
    \caption{R-S comparison of EKA.}
    \label{tab:RS_comparison}
    \small

    \resizebox{0.5\textwidth}{!}{\begin{tabular}{lcccccc|c}
    \toprule
             &\textbf{2Wiki} & \textbf{HotpotQA} & \textbf{Musique} & \textbf{NQ} & \textbf{PopQA} & \textbf{TriviaQA} & \textbf{Avg.}\\
        \midrule
        Graph-R1 &  55.24&  56.27&  52.95&  69.25&  61.55& 67.16& 60.40
\\
    \quad\quad\quad\quad+EKA &  \textbf{60.69}&  \textbf{60.36}& \textbf{61.54}&  \textbf{72.86}&  \textbf{64.97}& \textbf{68.99}& \textbf{64.90}
\\
        \bottomrule
    \end{tabular}}
\end{table}
We choose two RAG methods based on reinforcement learning as our backbone, Search-R1\citep{Search-R1} and Graph-R1\citep{Graph-R1}, accompanied with two different dataset splitting, to show our method's robustness across different methods and retrieval set. In Search-R1 setting, models are trained in two IND (in-domain) datasets (HotpotQA and NQ) and other datasets are OOD (out-of-domain) datasets for test. In Graph-R1 setting, models are trained within each dataset. Furthermore, a comprehensive retrieval set with chunks using the full Wikipedia corpus (Fullwiki) is used in the Search-R1 setting, and a smaller, dataset-specific structure-augmented retreival set is used in the Graph-R1 setting. We also run EKA on Search-R1 in the Graph-R1 setting with a smaller, dataset-specific chunk-based retreival set.
\subsection{Implementations}
\textbf{Baselines.} In Graph-R1 setting, we follow the previous work, including training-free methods from Graph-R1: NaiveGeneration, StandardRAG\citep{RAG}, GraphRAG\citep{GraphRAG}, LightRAG\citep{LightRAG}, PathRAG\citep{PathRAG}, HippoRAG2\citep{HippoRAG2}, HyperGraphRAG\citep{HyperGraphRAG} , training:SFT\citep{SFT}, R1\citep{GRPO}, R1-Searcher\citep{R1-Searcher} and Graph-R1\citep{Graph-R1} itself, we cite their performances for comparison if not specified. In the Search-R1 setting, additional baselines including CoT\citep{wei2022chain}, IRCoT\citep{trivedi2022interleaving}, Search-o1\citep{li2025search}, and Rejection Sampling\citep{ahn2024large} is compared. Detailed description of these baselines are put in the Appendix \ref{appendix:detailedimplementation}. We use Qwen2.5-7B-Instruct\citep{Qwen2.5} and Qwen2.5-14B-Instruct as LLM backbone for training. We also have done additional experiments on Qwen3\citep{Qwen3} in Appendix \ref{appendix:qwen3} and Section \ref{sec:experiments-trainingfree}.

\textbf{Retriever.} The retriever we used is highly dependent on the backbone. In Search-R1, the retriever is E5\citep{wang2022text}. In Graph-R1, the retriever is hypergraph-based retrieval with bge-large-en-v1.5\citep{BAAIembedding}.

\textbf{Datasets and Metrics.} Due to the different dataset splitting protocols in Search-R1 and Graph-R1, we conduct our experiments under both settings to ensure fair comparison. In Graph-R1 setting, we follow the original paper setting and use 6 common datasets\citep{FlashRAG} for QA, including 2Wikihop\citep{2WikiMultiHopQA}, HotpotQA\citep{HotpotQA}, Musique\citep{Musique}, NQ\citep{NQ}, PopQA\citep{PopQA}, TriviaQA\citep{TriviaQA}. Also in this setting we compare with Search-R1 baselines. We use EM, F1 and R-S to evaluate results. EM and F1 measures the answer and R-S measures the retrieval performances. In Search-R1 setting, we follow the original paper setting, appending one new dataset Bamboogle\citep{press2022measuring}, and using F1 score for comparison. Detailed information are referred to Appendix \ref{appendix:detailedimplementation}.
\begin{table*}[tbp]
    \centering
    \caption{Main results (F1 scores) compared in Search-R1 setting. The best performance is set in bold. $^\dagger/^\star$ represents IND/OOD datasets. Icons have the same meaning as Table \ref{tab:graphr1}.}\label{tab:searchr1}
    \scriptsize
    \setlength{\tabcolsep}{4pt}
    \renewcommand{\arraystretch}{1.2}

    \resizebox{\textwidth}{!}{\begin{tabular}{lccccccc|c}
        \toprule
        \textbf{Methods} & \multicolumn{3}{c}{\textbf{General QA}} & \multicolumn{4}{c}{\textbf{Multi-Hop QA}} \\
        \cmidrule{2-9}
         & NQ$^\dagger$ & TriviaQA$^\star$ & PopQA$^\star$ & HotpotQA$^\dagger$ & 2Wiki.$^\star$ & Musique$^\star$ & Bamboogle$^\star$ & Avg. \\
        \midrule
        \multicolumn{9}{c}{\textbf{\textit{Qwen2.5-7B-Instruct}}}  \\
        \raisebox{-0.22\height}{\includegraphics[width=0.02\textwidth]{close.png}}\raisebox{-0.22\height}{\includegraphics[width=0.02\textwidth]{none.png}} Direct Inference & 13.40 & 40.80 & 14.00 & 18.30 & 25.00 & 3.10 & 12.00 & 18.10 \\
       \raisebox{-0.22\height}{\includegraphics[width=0.02\textwidth]{close.png}}\raisebox{-0.22\height}{\includegraphics[width=0.02\textwidth]{none.png}} CoT & 4.80 & 18.50 & 5.40 & 9.20 & 11.10 & 2.20 & 23.20 & 10.60 \\
       \raisebox{-0.22\height}{\includegraphics[width=0.02\textwidth]{close.png}}\raisebox{-0.22\height}{\includegraphics[width=0.02\textwidth]{none.png}} IRCoT & 22.40 & 47.80 & 30.10 & 13.30 & 14.90 & 7.20 & 22.40 & 23.90 \\
        \raisebox{-0.22\height}{\includegraphics[width=0.02\textwidth]{close.png}}\raisebox{-0.22\height}{\includegraphics[width=0.02\textwidth]{chunk.png}} Standard RAG & 34.90 & 58.50 & 39.20 & 29.90 & 23.50 & 5.80 & 20.80 & 30.40 \\
        \midrule
       \raisebox{-0.22\height}{\includegraphics[width=0.02\textwidth]{open.png}}\raisebox{-0.22\height}{\includegraphics[width=0.02\textwidth]{none.png}} Search-o1 & 15.10 & 44.30 & 13.10 & 18.70 & 17.60 & 5.80 & 29.60 & 20.60 \\

       \raisebox{-0.22\height}{\includegraphics[width=0.02\textwidth]{open.png}}\raisebox{-0.22\height}{\includegraphics[width=0.02\textwidth]{none.png}}  SFT & 31.80 & 35.40 & 12.10 & 21.70 & 25.90 & 6.60 & 11.20 & 20.70  \\
       \raisebox{-0.22\height}{\includegraphics[width=0.02\textwidth]{open.png}}\raisebox{-0.22\height}{\includegraphics[width=0.02\textwidth]{none.png}}  R1-base & 29.70 & 53.90 & 20.20 & 24.20 & 27.30 & 8.30 & 29.60 & 27.60  \\
       \raisebox{-0.22\height}{\includegraphics[width=0.02\textwidth]{open.png}}\raisebox{-0.22\height}{\includegraphics[width=0.02\textwidth]{none.png}}  R1-instruct & 27.00 & 53.70 & 19.90 & 23.70 & 29.20 & 7.20 & 29.30 & 27.10  \\
        \raisebox{-0.22\height}{\includegraphics[width=0.02\textwidth]{open.png}}\raisebox{-0.22\height}{\includegraphics[width=0.02\textwidth]{chunk.png}} Rejection Sampling & 36.00 & 59.20 & 38.00 & 33.10 & 29.60 & 12.30 & 35.50 & 34.80 \\
       \rowcolor{Search-R1!15}\raisebox{-0.22\height}{\includegraphics[width=0.02\textwidth]{open.png}}\raisebox{-0.22\height}{\includegraphics[width=0.02\textwidth]{chunk.png}}  Search-R1 & 39.30 & 61.00 & 39.70 & 37.00 & 41.40 & 14.60 & 36.80 & 38.50 \\
       \rowcolor{Search-R1!15} \raisebox{-0.22\height}{\includegraphics[width=0.02\textwidth]{open.png}}\raisebox{-0.22\height}{\includegraphics[width=0.02\textwidth]{chunk.png}}  \quad\quad\quad\quad\quad +EKA& \textbf{49.80} & \textbf{66.10} & \textbf{48.90} & \textbf{45.70} & \textbf{42.70} & \textbf{20.90} & \textbf{39.50} & \textbf{44.80} \\
       \rowcolor{Search-R1!15} \raisebox{-0.22\height}{\includegraphics[width=0.02\textwidth]{open.png}}\raisebox{-0.22\height}{\includegraphics[width=0.02\textwidth]{chunk.png}}  \quad\quad\quad\quad\quad $\Delta$ & \emph{+10.50} & \emph{+5.10} & \emph{+9.20} & \emph{+8.70} & \emph{+1.30} & \emph{+6.30} & \emph{+2.70} & \emph{+6.30} \\
        \bottomrule
        \end{tabular}}
\end{table*}
\subsection{Comparison in Graph-R1 Setting}
We show the results in Table \ref{tab:graphr1}. Note that Search-R1 uses PPO method in its paper but Graph-R1 runs GRPO in their experiments so we run the Search-R1-PPO by ourselves as the PPO variants in the table. We found that EKA improves the performance of Graph-R1 by an average of 3 F1 points, Search-R1 by an average of 11 F1 points and Search-R1-PPO by an average of 7 F1 points, demonstrating a substantial performance gain across different RL methods. Also, the improvement in R-S scores indicates that EKA can actually improve the exploitation in focusing retrieval necessary information.

Then we analysis the R-S of EKA compared with Graph-R1 in Table \ref{tab:RS_comparison}. This suggests that EKA's performance gains are partially driven by improved retrieval quality.

\subsection{Comparison in Search-R1 Setting}

In Search-R1 setting, we show the results of using Fullwiki as the retrieval set to show our methods' robustness in retrieval set. As constructing a full Wikipedia hypergraph in the manner of Graph-R1 is currently computationally prohibitive, we only use Search-R1 as our backbone. The results shows that EKA also can increase performances when the retrieval set is very large, and can show incremental performances in both IND and OOD datasets in Table \ref{tab:searchr1}. Notably, EKA improves the performance of Search-R1 by an average of 6.3 F1 points.

\subsection{Training-free EKA}
\label{sec:experiments-trainingfree}
To demonstrate versatility and scalability, we evaluate EKA as a \textit{training-free} inference module on larger models where RL fine-tuning is computationally prohibitive. By aligning with the retrieval set before reasoning, EKA consistently delivers substantial gains across benchmarks (Table~\ref{tab:training_free}). These results confirm that "plan failure" from ungrounded thinking persists even in large-scale models, and EKA serves as a robust, plug-and-play solution to mitigate hallucinations and enhance reasoning stability without parameter updates.

\begin{table}[h]
    \centering
    \small
    \setlength{\tabcolsep}{3.5pt}
    \caption{Performance (F1 Score) of EKA as a training-free inference strategy on large-scale models. EKA consistently improves performance across all datasets without any parameter updates.}
    \resizebox{0.5\textwidth}{!}{\begin{tabular}{lcccccc}
         \toprule
         \textbf{Model} & \textbf{2Wiki} & \textbf{HotpotQA} & \textbf{Musique} & \textbf{NQ} & \textbf{PopQA} & \textbf{TriviaQA} \\
         \midrule
         Qwen2.5-32B-Instruct & 13.73 & 23.96 & 8.29 & 11.62 & 15.19 & 23.65 \\
         \quad + EKA & \textbf{18.17} & \textbf{26.14} & \textbf{13.04} & \textbf{15.63} & \textbf{17.08} & \textbf{27.84} \\
         \midrule
         Qwen3-235-A30B-Instruct & 30.56 & 37.80 & 19.93 & 21.49 & 28.73 & 38.55 \\
         \quad + EKA & \textbf{38.39} & \textbf{48.82} & \textbf{28.17} & \textbf{24.68} & \textbf{33.61} & \textbf{44.72} \\
         \bottomrule
    \end{tabular}}
    
    \label{tab:training_free}
\end{table}

\section{Ablations}

Experiments are done in the Graph-R1 setting in the ablation section, and we aim to answer the following three questions:
\begin{itemize}
    \item Q1. Why Early Knowledge Alignment can make the performance better, from an entropy perspective.
    \item Q2. Can Early Knowledge shorten the number of thinking turns? And what is metrics' dynamics in every step in the training?
    \item Q3. Will Early Knowledge Alignment in RL training downgrade the generalization of trained models? 
\end{itemize}

\subsection{Entropy Analysis}
\label{ablation:entropy}

\begin{figure}[htbp]
    \centering
    \subfigure[]{
        \includegraphics[width=0.13\textwidth]{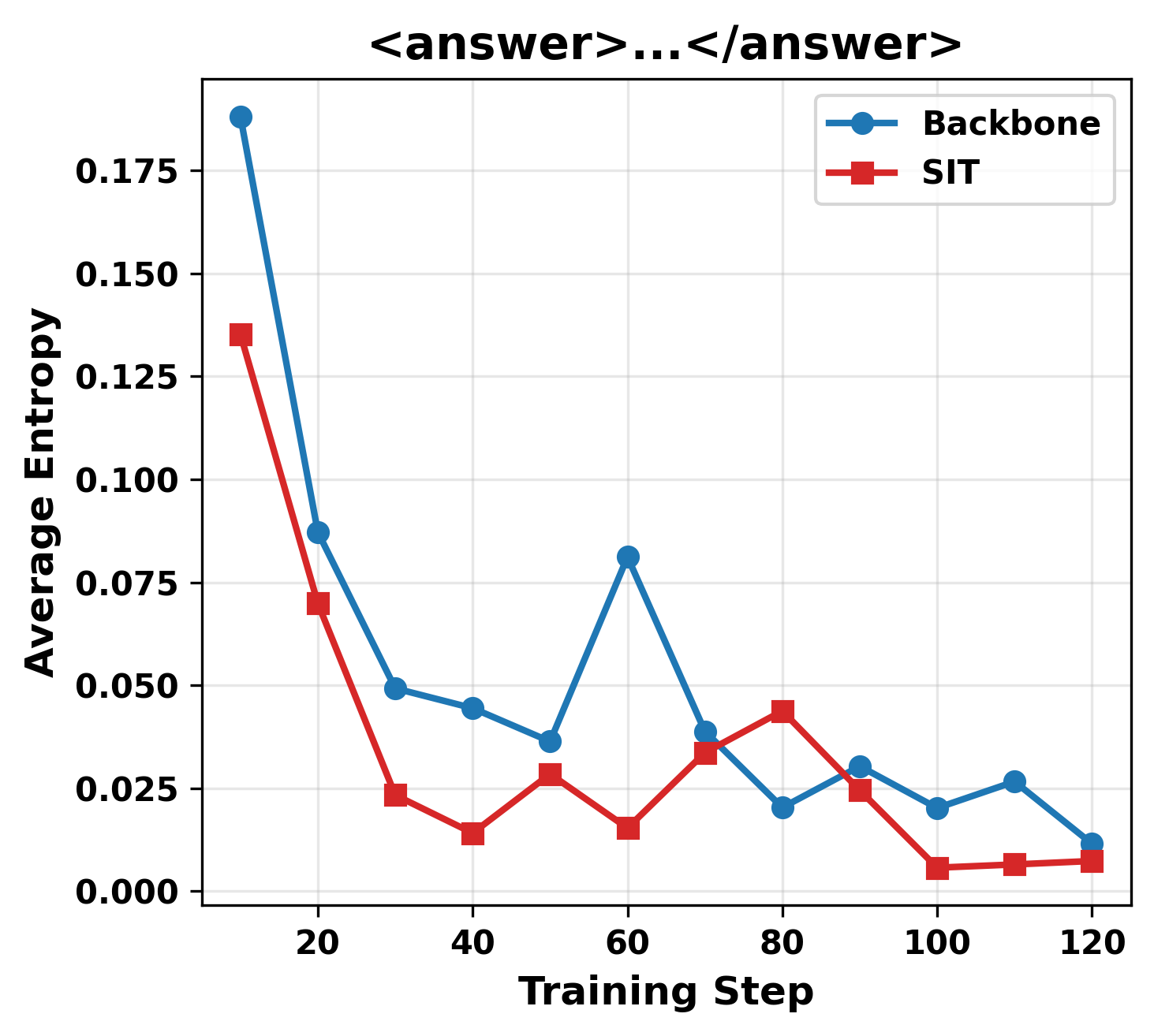}
    }
    \subfigure[]{
        \includegraphics[width=0.13\textwidth]{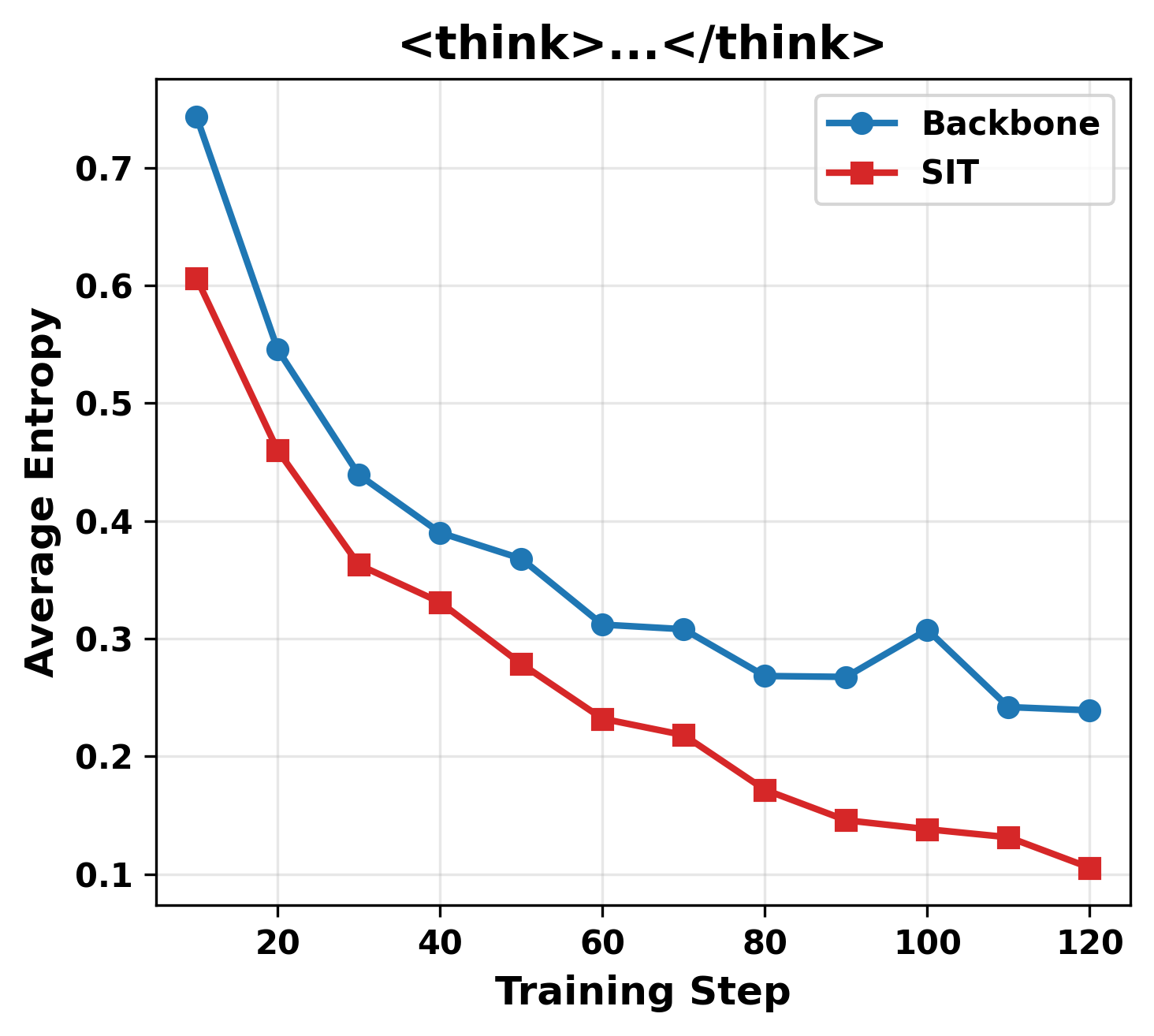}
    }
    \subfigure[]{
        \includegraphics[width=0.13\textwidth]{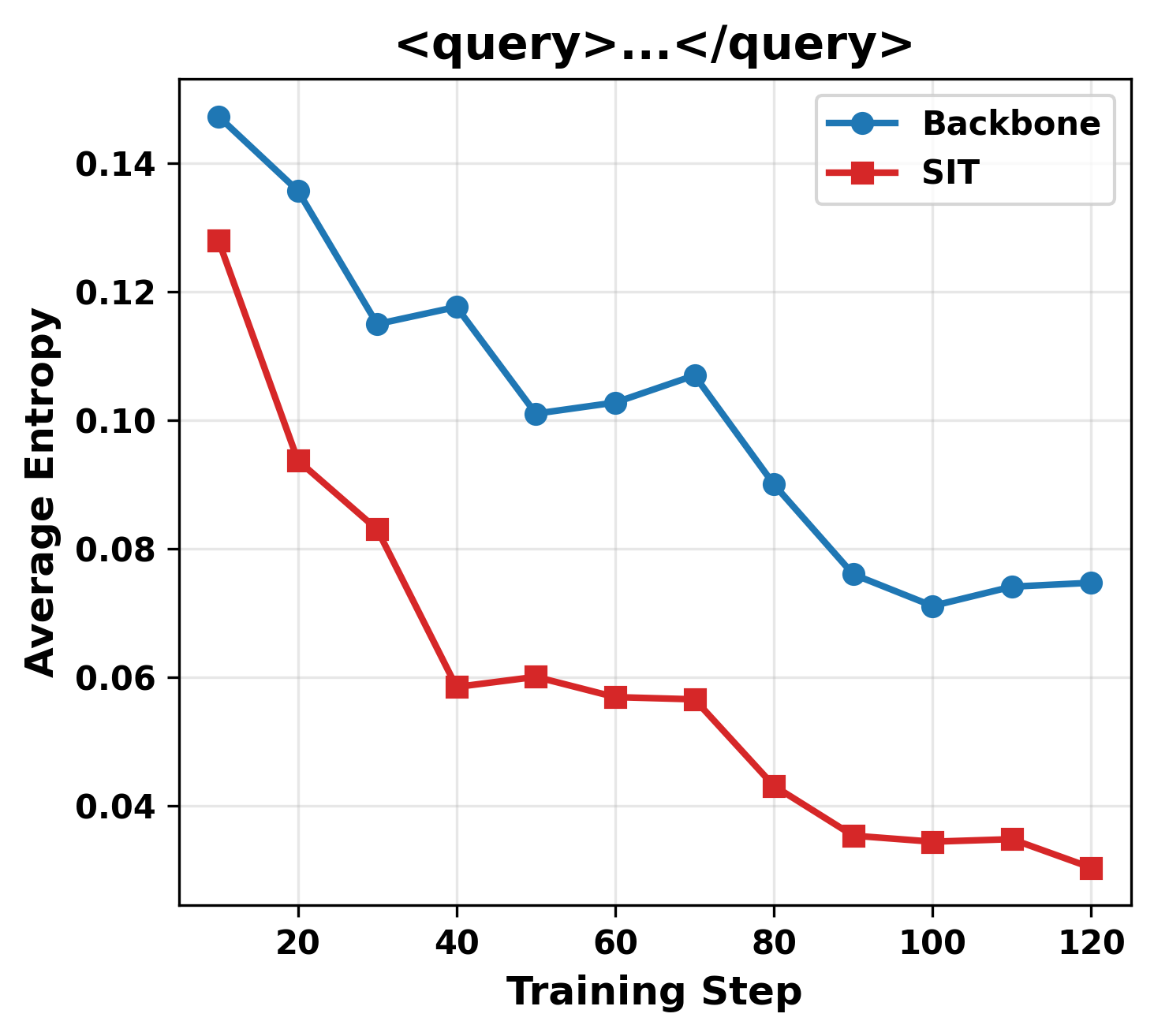}
    }
    \caption{Entropy comparison of backbone (Graph-R1) and EKA. (a), (b), and (c) show average entropy of tokens between "<answer>...</answer>", "<think>...</think>", "<query>...</query>".}
    \label{fig:entropy}
\end{figure}

In RL training, the entropy demonstrates model's exploration ability in training. However, in the context of multi-hop RAG, unconstrained exploration is not always beneficial, as the reasoning process must remain aligned with the information available in the retrieval set. EKA is designed precisely to provide this initial alignment. We show the comparison of Graph-R1's entropy of tokens between "<answer>...</answer>", "<think>...</think>", "<query>...</query>" with EKA or not in Figure \ref{fig:entropy}.

We found that the entropy values for all action types are generally lower with EKA than without it. At zero step with the same LLM, the lower entropy of tokens between "<answer>" "</answer>" (which is actually the answer tokens) of EKA fits the intermediate conclusion in the proof in Appendix \ref{appendix:theoreticalproof} that 
\begin{equation}
    \mathbb{E}_\pi \left[ I(A^\star; \mathcal H_T^{EKA} \mid Q) \right] \ge  \mathbb{E}_\pi \left[ I(A^\star; \mathcal H_T \mid Q) \right],
\end{equation}
which predicts the lower entropy of EKA answer tokens. Although there is a single training step where the answer entropy for EKA is momentarily higher, the overarching trend shows that EKA consistently leads to lower answer token entropy.

Besides, the lower entropy of think and search tokens show that LLM with EKA has more determined exploration direction in thinking and searching, which is exactly what we assume in the beginning.

\subsection{Shorter turns and Metrics Dynamics.}

We show that with EKA, the exploration turns of LLMs shrinks about one turn on average in Table \ref{tab:ablation_turns}. Shorter turns means less noise in the retrieval that can make LLM more focus on the right information.

\begin{figure}[htbp]
    \centering
    \subfigure[]{
        \includegraphics[width=0.13\textwidth]{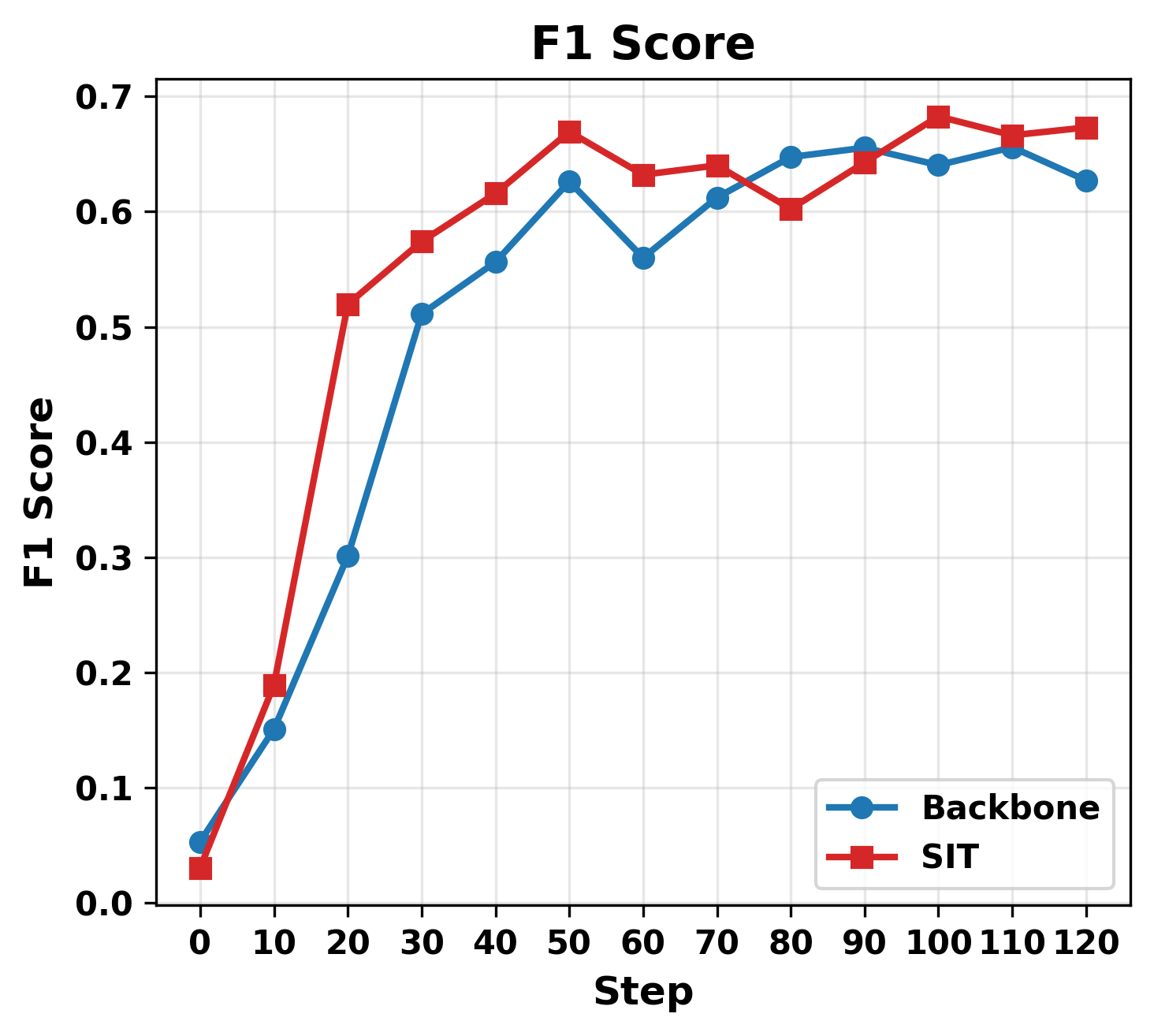}
    }
    \subfigure[]{
        \includegraphics[width=0.13\textwidth]{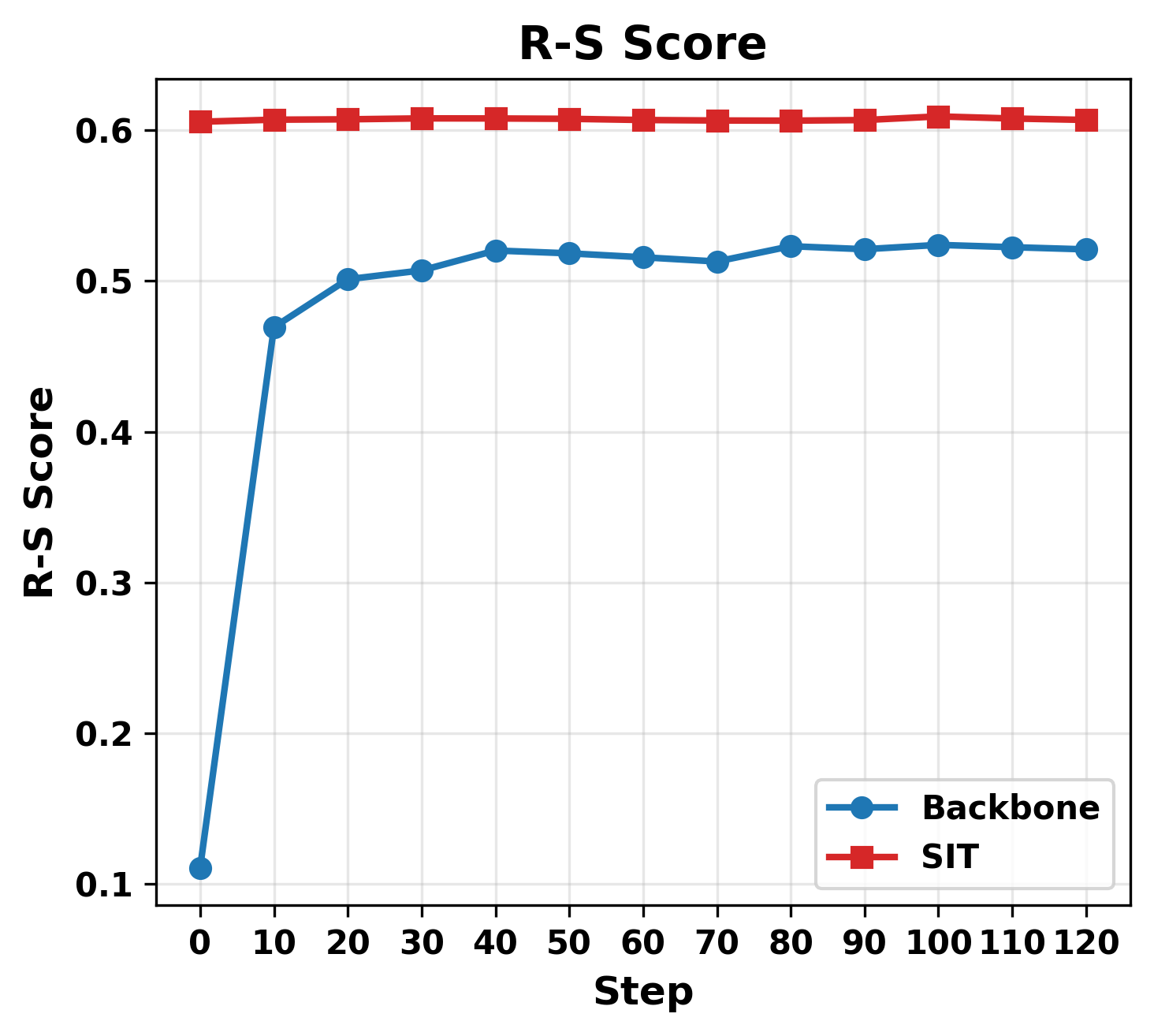}
    }
    \subfigure[]{
        \includegraphics[width=0.13\textwidth]{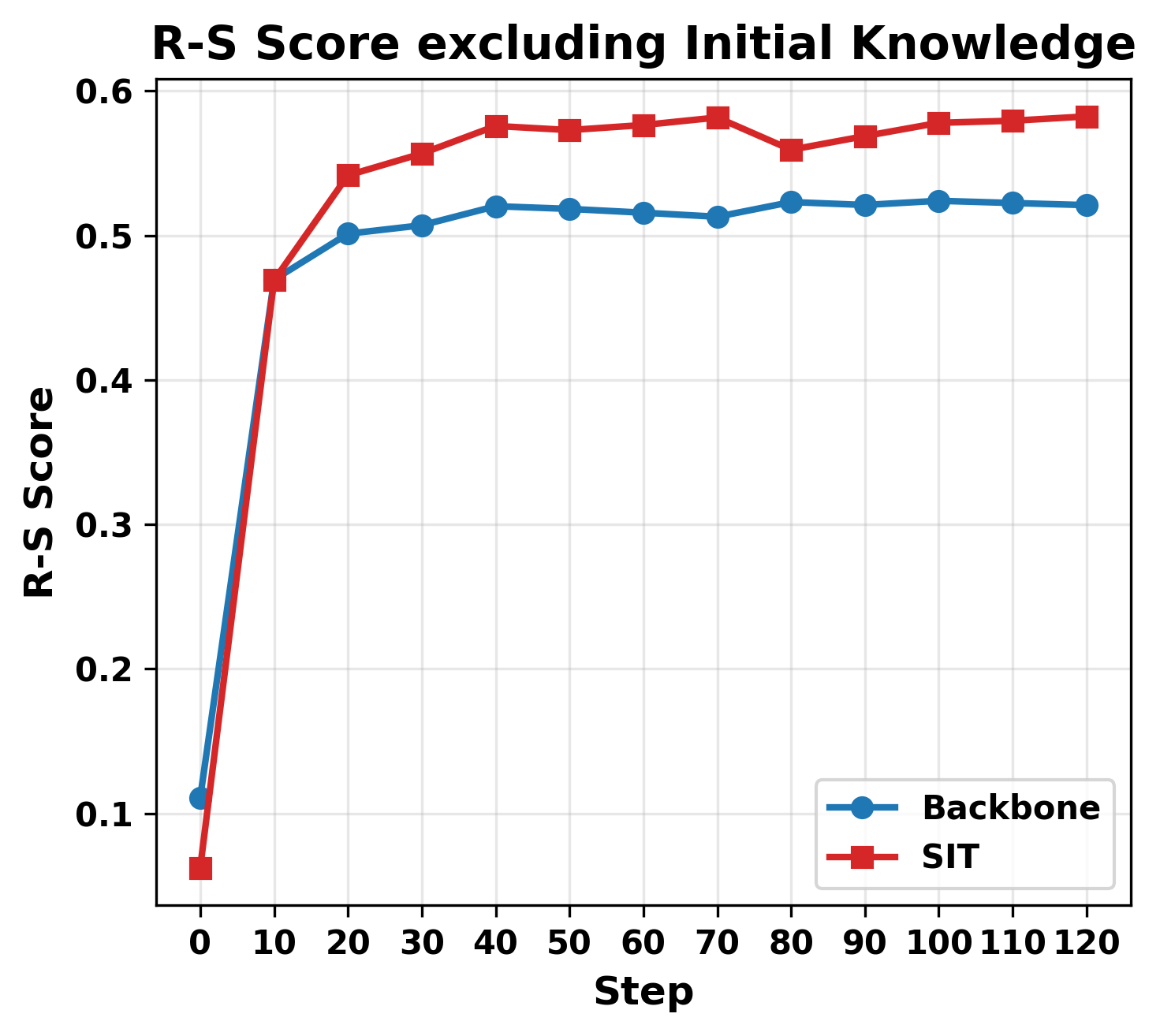}
    }
    \caption{F1 and R-S scores per training step on the 2Wiki dataset. (a) F1 score. (b) R-S score. (c) R-S score excluding the early knowledge. }
    \label{fig:ablation_F1_steps}
\end{figure}

\begin{table}[htbp]
    \centering
    \small
    \caption{Average turns of Graph-R1 with or without EKA.}
    \label{tab:ablation_turns}
    \resizebox{0.5\textwidth}{!}{\begin{tabular}{lcccccc|c}
    \toprule
             &\textbf{2Wiki} & \textbf{HotpotQA} & \textbf{Musique} & \textbf{NQ} & \textbf{PopQA} & \textbf{TriviaQA} & \textbf{Avg.}\\
        \midrule
        Graph-R1 &  3.12&  3.12&  3.88&  3.06&  3.53& 2.82& 
3.26\\
        \quad\quad+EKA &  \textbf{2.72}&  \textbf{2.80}&  \textbf{2.68}&  \textbf{1.52}&  \textbf{1.91}& \textbf{1.72}& \textbf{2.22}\\
        \bottomrule
    \end{tabular}}
\end{table}
Next, we show the F1 and R-S scores in the training step in Figure \ref{fig:ablation_F1_steps}. We found that with EKA, our model's RS is high from the beginning. Even when we exclude the early knowledge in computing the metrics, the R-S score of backbone with EKA can still increase to a higher value than the model without EKA.

\subsection{Generalization}
\subsubsection{Generalization across datasets}
While the generalization performance on OOD datasets using the Search-R1 backbone was presented in Table \ref{tab:searchr1}, this section evaluates the generalization of EKA with the Graph-R1 backbone. The results show that our method not only achieves better results in IID conditions but also show better generalization results on average than without EKA. 
\begin{table*}[htbp]
    \centering
    \caption{Generalization test on backbone and EKA. The row datasets are training datasets and the column datasets are test datasets.}
    \label{tab:generalization_experiments}
    \small
    
    \begin{tabular}{lcccccc|c}
    \toprule
        Train Datasets &2Wiki. & HotpotQA & Musique & NQ & PopQA & TriviaQA & Avg. \\
        \midrule
        2Wiki. &  65.04 &  59.92 &  35.92 &  45.24 &  42.57 & 63.38 & 52.01 
\\
        \quad\quad+EKA &  \textbf{68.26} &  63.90 &  44.53 &  46.89 &  50.78 & 65.53 & \textbf{56.65} 
\\
 \quad\quad+$\Delta$&  \emph{+3.22} &  \emph{+3.98} &  \emph{+8.61} &  \emph{+1.65} &  \emph{+8.21} & \emph{+2.15} & \emph{+4.64} 
\\
        \midrule
       HotpotQA  &  58.27 &  62.69 &  33.27 &  37.89 &  44.30 & 57.20 & 48.94 
\\
       \quad\quad+EKA &  60.86 &  \textbf{66.14} &  38.87 &  45.14 &  47.60 & 66.96 & 54.26 
\\
 \quad\quad+$\Delta$&  \emph{+2.59} &  \emph{+3.45} &  \emph{+5.60} &  \emph{+7.25} &  \emph{+3.30} & \emph{+9.76} & \emph{+5.32} 
\\
       \midrule
       Musique  &  43.87 &  52.32 &  46.17 &  43.66 &  44.76 & 64.45 & 49.21 
\\
       \quad\quad+EKA  &  54.90 &  59.99 &  \textbf{51.63} &  47.63 &  48.98 & 69.82 & 55.49 
\\
 \quad\quad+$\Delta$&  \emph{+11.03} &  \emph{+7.67} &  \emph{+5.46} &  \emph{+3.97} &  \emph{+4.22} & \emph{+5.37} & \emph{+6.28} 
\\
       \midrule
        NQ &  52.13 &  53.19 &  34.57 &  49.87 &  43.10 & 63.74 & 49.43 
\\
        \quad\quad+EKA &  54.77 &  55.83 &  37.75 &  \textbf{51.99} &  48.72 & 67.38 & 52.74 
\\
 \quad\quad+$\Delta$&  \emph{+2.64} &  \emph{+2.64} &  \emph{+3.18} &  \emph{+2.12} &  \emph{+5.62} & \emph{+3.64} & \emph{+3.31} 
\\
        \midrule
       PopQA  &  47.41 &  58.45 &  35.99 &  43.40 &  51.22 & 68.91 & 50.90 
\\
       \quad\quad+EKA &  48.51 &  57.52 &  34.66 &  43.88 &  \textbf{53.49} & 69.98 & 51.34 
\\
 \quad\quad+$\Delta$&  \emph{+1.10} &  \emph{-0.93}&  \emph{-1.33}&  \emph{+0.48} &  \emph{+2.27} & \emph{+1.07} & \emph{+0.44} 
\\
       \midrule
        TriviaQA &  46.83 &  53.82 &  22.87 &  41.66 &  44.71 & 71.93 & 46.97 
\\
        \quad\quad+EKA &  52.17 &  55.18 &  31.31 &  44.87 &  47.23 & \textbf{72.37} & 50.52 
\\
 \quad\quad+$\Delta$&  \emph{+5.34} &  \emph{+1.36} &  \emph{+8.44} &  \emph{+3.21} &  \emph{+2.52} & \emph{+0.44} & \emph{+3.55} 
\\
        \bottomrule
    \end{tabular}
\end{table*}

\subsubsection{Mismatched Early Knowledge}

We further investigate the robustness of Early Knowledge Alignment (EKA) against variations in the quality and source of the early knowledge $P_0$.

\textbf{Noisy Early Knowledge.}
In real-world scenarios, the Early Knowledge $P_0$ may contain irrelevant information or noise. To simulate this, we conduct experiments using the full Wikipedia corpus as the retrieval source for the initial step (denoted as EKA-wiki), which introduces significantly more noise compared to the dataset-specific retrieval sets. 
As shown in Table~\ref{tab:ablation-noiseik}, although the introduction of noise in EKA-wiki leads to a slight performance drop compared to the standard EKA, it still consistently outperforms the baseline without EKA in average. This demonstrates that the benefit of EKA comes from the \textit{grounding} effect of the early knowledge, which remains effective even when it is imperfect.
\begin{table}[h]
    \centering
    \small
    \setlength{\tabcolsep}{4pt}
        \caption{Performance(F1 Score) comparison with noisy early knowledge.}
    \resizebox{0.5\textwidth}{!}{\begin{tabular}{lcccccc}
         \toprule
         \textbf{Method} & \textbf{2Wiki} & \textbf{HotpotQA} & \textbf{Musique} & \textbf{NQ} & \textbf{PopQA} & \textbf{TriviaQA} \\
         \midrule
         Qwen2.5-7B-Instruct & 65.04 & 62.69 & 46.17 & 49.87 & 51.22 & 71.93 \\
         + EKA (Standard) & \textbf{68.26} & \textbf{66.14} & \textbf{51.63} & \textbf{51.99} & 53.49 & \textbf{72.37} \\
         + EKA-wiki (Noisy) & 66.18 & 62.91 & 47.16 & 50.43 & \textbf{53.98} & 71.77 \\
         \bottomrule
    \end{tabular}}

    \label{tab:ablation-noiseik}
\end{table}

\textbf{Mismatched Retriever.}
To verify that our improvements are not dependent on a specific retrieval model, we evaluate EKA using different dense retrievers. We compare the default BGE retriever (EKA-bge) with the E5 retriever (EKA-e5). 
Table~\ref{tab:ablation-retriever} presents the results across six datasets. We observe that EKA yields consistent performance gains regardless of the retriever used, confirming that the EKA framework is retriever-agnostic and generalizes well across different semantic embedding spaces.

\begin{table}[H]
    \centering
    \small
    \setlength{\tabcolsep}{4pt}
        \caption{Ablation study on retriever quality.}
    \resizebox{0.5\textwidth}{!}{\begin{tabular}{lcccccc}
         \toprule
         \textbf{Method} & \textbf{2Wiki} & \textbf{HotpotQA} & \textbf{Musique} & \textbf{NQ} & \textbf{PopQA} & \textbf{TriviaQA} \\
         \midrule
         Qwen2.5-7B-Instruct & 65.04 & 62.69 & 46.17 & 49.87 & 51.22 & 71.93 \\
          EKA-bge (Standard) & \textbf{68.26} & \textbf{66.14} & 51.63 & \textbf{51.99} & \textbf{53.49} & \textbf{72.37} \\
         EKA-e5 & 68.18 & 64.74 & \textbf{54.27} & 50.74 & 53.46 & 72.21 \\

         \bottomrule
    \end{tabular}}
    \label{tab:ablation-retriever}
\end{table}
\vspace{-3mm}
\section{Conclusion}
\vspace{-3mm}
All in all, we propose an easy but effective module in iterative RAG pipeline called Early Knowledge Alignment (EKA) that can guide right directions of thinking, resulting in more efficient exploration in RL training and better end-to-end performances. Our comprehensive experiments rigorously validate the efficacy and robustness of EKA. The approach delivers substantial performance gains to state-of-the-art RL-based frameworks, including Search-R1 and Graph-R1, across diverse RL algorithms (PPO and GRPO) and varied retrieval contexts—from small, structured corpora to large-scale, unstructured document sets. In addition, EKA consistently maintains or even improves upon the generalization capabilities of the backbone models, showcasing its reliability. Crucially, we also demonstrate EKA's scalability as a plug-and-play, training-free module for large models. This motivates us the shift of designing advanced RAG systems: from a plan-first model to the early knowledge alignment process.
\section{Limitations}
While Early Knowledge Alignment achieves performances in multi-hop QA, whether it works in much more complex Deepresearch scenerios remains undiscovered.

\section{Reproducibility Statement}
We present a detailed training algorithm in Appendix \ref{sec:appendix-algo}, technical proofs in Appendix \ref{appendix:theoreticalproof}, and additional experimental/implementation details in Appendix \ref{appendix:detailedimplementation}. Additionally, code for our model is uploaded as supplemental materials with the submission.


\bibliography{custom}

\appendix
\onecolumn


\section{Algorithm}
\label{sec:appendix-algo}

\begin{algorithm}[htbp]
\caption{Early Knowledge Alignment}
\label{alg:llm_search}
\begin{algorithmic}[1]
\Require Input \( x \), LLM \( \pi_{\theta} \), Retrieval set \( \mathcal{R} \), Max turns \( B \).
\Ensure Output \( y \).

\State Initialize \( y \gets \emptyset \)
\State Initialize \( b \gets 0 \)
\State Initialize Searching Knowledge $\mathcal{P}_0= \mathcal{R}(x)$ and update \( x \gets x + \mathcal{P}_0 \)

\While{\( b < B \)}
    \State Rollout   \( y_b \gets \emptyset \) 
    \While{True}
    \State Generating \( y_t \sim \pi_{\theta}(\cdot \mid x, y + y_b) \)
    \State concatenate token   \( y_b \gets y_b + y_t \)
    \If{\( y_t \) in [\textcolor{orange}{\texttt{</query>}}, \textcolor{purple}{\texttt{</answer>}}, \texttt{<eos>}]}
        break
    \EndIf
    \EndWhile

    \State \( y  \gets  y + y_b \)
    \If{extract \textcolor{orange}{\texttt{<query>}} \textcolor{orange}{\texttt{</query>}} from \( y_b \)}
        \State Extract \( q \gets \text{Parse}(y_b,\textcolor{orange}{\texttt{<query>}}, \textcolor{orange}{\texttt{</query>}} ) \)
        \State Retrive knowledge \( d = \mathcal{R}(q) \)
        \State Continue rollout \( y  \gets  y + \textcolor{teal}{\texttt{</knowledge>}}d\textcolor{teal}{\texttt{</knowledge>}}  \)
    \ElsIf{extract \textcolor{purple}{\texttt{</answer>}} from \( y_b \)}
        \State \textbf{return} \( y \)
    \EndIf

    \State count turns \( b \gets b + 1 \)
\EndWhile

\State \textbf{return} \( y \)
\end{algorithmic}
\end{algorithm}

\section{Additional Experiments}
\subsection{Qwen3 Model Results}
\label{appendix:qwen3}
We show the Qwen3-4B-Instruct-2507 model's performances in the training step in Figure \ref{appendix:qwen3}. It is shown that even bad results, EKA can still improve Qwen3 performances. We check the output of Qwen3 and find that the reason is that Qwen3 instruction models have used "think" token in its pre-train so when they have removed think pattern in 2507 model, it's hard for the model to generate the thinking process in the pipeline, resulting in low performances.
\begin{figure}
    \centering
    \includegraphics[width=0.5\linewidth]{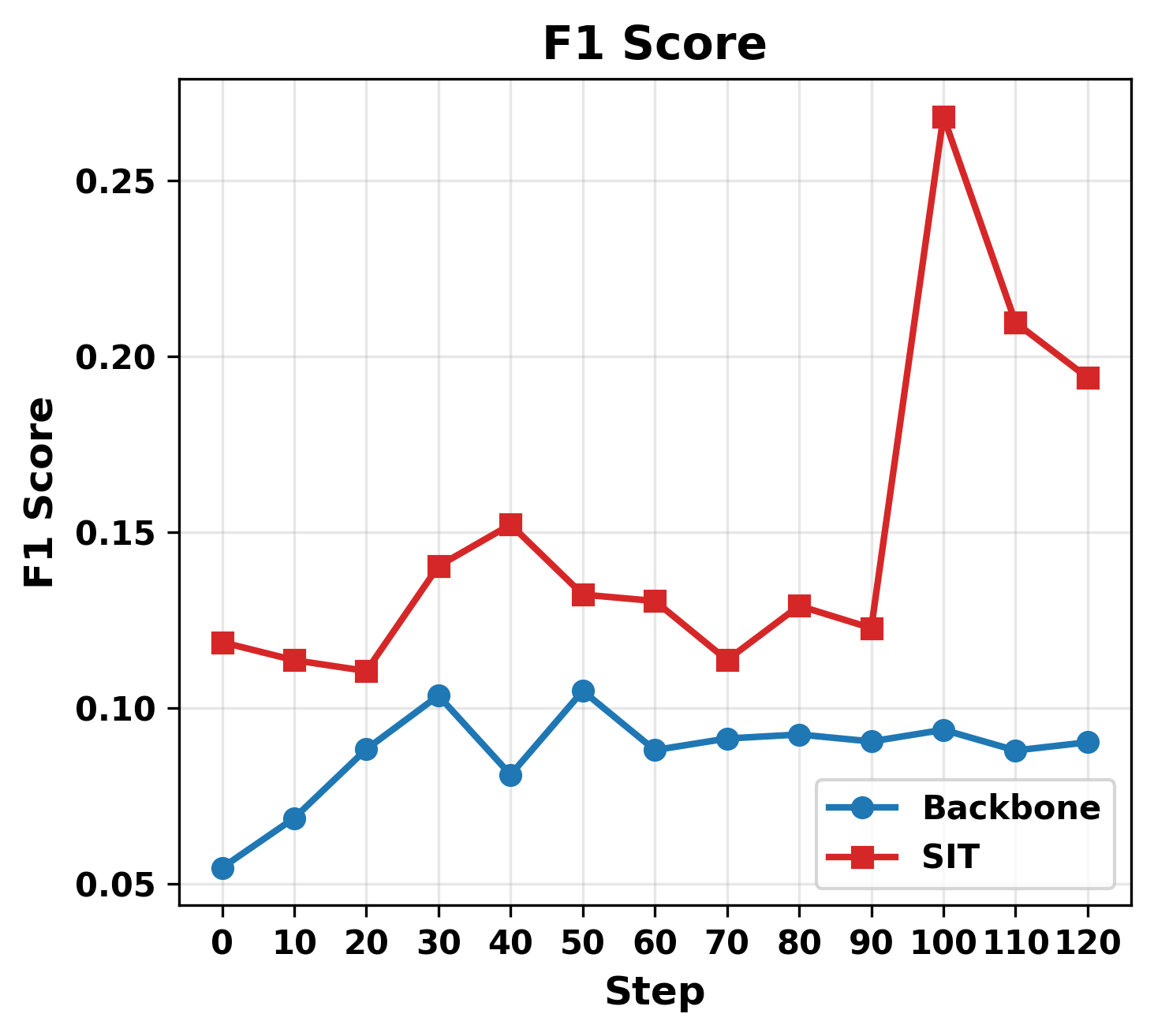}
    \caption{Qwen3-4B-Instruct-2507 model's F1 score in each step in 2Wiki dataset. Backbone is Graph-R1.}
    \label{fig:appendix-qwen3}
\end{figure}

\subsection{Case Study}
\label{appendix:casestudy}
In this section, we show a classical example of why Early Knowledge Alignment is useful. In Graph-R1, when the model lacks planning ability to split the question into two parts, it will generate a useless searching for both two things in turns and turns. As shown in Table \ref{tab:case-graphr1}, it fails to retrieve the directors. While as shown in Table \ref{tab:case-graphr1+EKA} the model with EKA knows searching for two things is useless, then it will split the question and search for two directors separately and finally retrieve the right documents, resulting in the right answer.

\begin{table}[h]
    \centering
    \footnotesize
    \caption{A case study of Graph-R1.}\label{tab:case-graphr1}
    \begin{tabular}{p{13.5cm}}
        \hline
        \textbf{Question}: Which film has the director born later, I'Ll Tell The World or Saranggola?\\
        \hline
        \textbf{Ground Truth}: ``Saranggola''\\
        \hline
        \textbf{Graph-R1}: \\
        \think{To answer this question, I need to find information about the birth years of the directors of the movies I'Ll Tell The World and Saranggola, then compare them.} \\
        \search{SELECT directorBirthYear WHERE movieName = "I'Ll Tell The World" OR movieName = "Saranggola"}\\
        \info{Invalid tool call format. Please use <query>\{ "query": "statement" \}</query> format.}\\
        \think{The previous query did not work as expected, so I'll try a different approach. First, I need to get the birth years of the directors of both movies.}\\
        \search{SELECT directorBirthYear WHERE movieName = "I'Ll Tell The World" OR movieName = "Saranggola"}\\
        \info{Invalid tool call format. Please use <query>\{ "query": "statement" \}</query> format.}\\
        \think{Given the error in my previous attempts, I'll directly query for the birth years of the directors of I'Ll Tell The World and Saranggola, and then compare them.}\\
        \search{\{ "query": "SELECT directorBirthYear WHERE movieName = 'I'Ll Tell The World' OR movieName = 'Saranggola'" \}}\\
        \info{\{ "results": [
            \{ "<knowledge>": "Lena Dunham (born May 13, 1986) is an American actress, writer, director, and producer.", "<coherence>": 1.0 \},
            \{ "<knowledge>": "Priyadarshan (born January 30, 1957) is an Indian film director, producer, and screenwriter.", "<coherence>": 0.508 \},
            \{ "<knowledge>": "V. S. Raghavan (director)", "<coherence>": 0.333 \}
        ] \}}\\
        \think{I found the birth years of the directors of both movies. Lena Dunham is the director of I'Ll Tell The World and was born in 1986. While V. S. Raghavan is the director of Saranggola and was born in 1935. Since Lena Dunham's birth year is later than V. S. Raghavan, the film I'Ll Tell The World has the director born later.}\\
        \answer{I'Ll Tell The World} \\
        \hline
    \end{tabular}
\end{table}

\begin{table}[h]
    \centering
    \footnotesize
    \caption{A case study of Graph-R1+EKA.}\label{tab:case-graphr1+EKA}
    \begin{tabular}{p{13.5cm}}
        \hline
        \textbf{Question}: Which film has the director born later, I'll Tell The World or Saranggola?\\
        \hline
        \textbf{Ground Truth}: ``Saranggola''\\
        \hline
        \textbf{Graph-R1+EKA}: \\
        \think{From the knowledge provided, I'll Tell The World is a 1945 American comedy film directed by Leslie Goodwins. Saranggola is a 1999 Filipino drama film directed by Gil Portes. To determine which film has the director born later, I will need to find the birth years of Leslie Goodwins and Gil Portes.}\\
        \search{\{ ``type'': ``search'', ``query'': ``Leslie Goodwins birth year'' \}}\\
        \info{\{``results'': [\{``<knowledge>'': ``They Met in Argentina is a 1941 American film directed by Leslie Goodwins and Jack Hively for RKO Pictures.'', ``<coherence>'': 1.7\}, \{``<knowledge>'': ``Leslie Goodwins'', ``<coherence>'': 1.0\}, \{``<knowledge>'': ``Leslie Goodwins (17 September 1899 -- 8 January 1969) was an English film director and screenwriter.'', ``<coherence>'': 0.819\}, \{``<knowledge>'': ``With Love and Kisses is a 1936 American musical comedy film directed by Leslie Goodwins and starring Pinky Tomlin, Toby Wing and Kane Richmond.'', ``<coherence>'': 0.643\}, \{``<knowledge>'': ``I'll Tell the World is a 1945 American comedy film directed by Leslie Goodwins and written by Henry Blankfort and Lester Pine.'', ``<coherence>'': 0.333\}, \{``<knowledge>'': ``The Mexican Spitfire's Baby is a 1941 American comedy film directed by Leslie Goodwins.'', ``<coherence>'': 0.25\}, \{``<knowledge>'': ``Leslie Goodwins is acknowledged but lacks detailed background in the provided text.'', ``<coherence>'': 0.2\}, \{``<knowledge>'': ``Murder in the Blue Room is a 1944 American film directed by Leslie Goodwins. It is the second remake of the 1933 American Pre-Code murder-mystery film `Secret of the Blue Room.''', ``<coherence>'': 0.167\}]\}}\\
        \think{From the knowledge provided, Leslie Goodwins was born on September 17, 1899. Now I need to find the birth year of Gil Portes to determine which film has the director born later.}\\
        \search{\{``type'': ``search'', ``query'': ``Gil Portes birth year''\}}\\
        \info{\{``results'': [\{``<knowledge>'': ``Gil M. Portes (September 13, 1945 -- May 24, 2017) was a Filipino film director, film producer and screenwriter.'', ``<coherence>'': 1.125\}, \{``<knowledge>'': ``Saturday Night (1950 film) is a 1950 Spanish drama film directed by Rafael Gil.'', ``<coherence>'': 1.0\}, \{``<knowledge>'': ``The Man Who Wanted to Kill Himself (1970 film)'', ``<coherence>'': 0.5\}, \{``<knowledge>'': ``Carlos Gardel (born Charles Romuald Gardès; 11 December 1890 -- 24 June 1935) was a French Argentine singer, songwriter, composer and actor.'', ``<coherence>'': 0.5\}, \{``<knowledge>'': ``Rafael Gil was a Spanish film director and screenwriter.'', ``<coherence>'': 0.4\}, \{``<knowledge>'': ``The Autonomines is a Spanish comedy film directed by Rafael Gil and starring Alfredo Landa, María Casanova, and Manolo Codeso.'', ``<coherence>'': 0.333\}, \{``<knowledge>'': ``Gil was a prominent director of the Franco era.'', ``<coherence>'': 0.333\}, \{``<knowledge>'': ``El canto del gallo is a 1955 Spanish drama film directed by Rafael Gil.'', ``<coherence>'': 0.25\}, \{``<knowledge>'': ``Gil's film `La noche del sábado' was nominated for the Gold Lion at the 1950 Venice Film Festival.'', ``<coherence>'': 0.25\}, \{``<knowledge>'': ``Saranggola (international title: The Kite) is a 1999 Filipino drama film directed by Gil Portes, starring Ricky Davao, Lester Llansang and Jennifer Sevilla.'', ``<coherence>'': 0.167\}]\}}\\
        \think{From the knowledge provided, Leslie Goodwins was born on September 17, 1899. Gil Portes was born on September 13, 1945. Now I can compare the birth years to determine which film has the director born later.}\\
        \answer{Saranggola}\\
        \hline
    \end{tabular}
\end{table}

\section{Theoretical Proof}
\label{appendix:theoreticalproof}

\textbf{Proposition 1.} Early Knowledge Alignment is better than traditional thinking in iterative RAG from an entropy perspective.
\begin{proof}
Given the condition of iterative RAG for an LLM $\pi$ divides the budget across $T$ rounds as $B = \sum_{t=1}^T B_t$. At each round $t>=1$, we denote $\mathcal P_t$ as the retrieval results at this step, and the prior evidence $\mathcal H_{t-1} = \{\mathcal P_1, \dots, \mathcal P_{t-1}\}$. The LLM uses $\mathcal H_{t-1}$ to update its internal belief $h_{t-1}$ and selects new evidence $\mathcal P_t$ of size $B_t$ by actively exploring the graph based on current uncertainty. The updated belief $h_t$ is obtained via Bayesian inference, and the entire process forms a dynamic system:
\begin{equation}
h_t = f(h_{t-1}, \mathcal P_t, R_G).
\end{equation}
To evaluate retrieval progress, we define a Lyapunov-style potential function $V_t = H(A^\star \mid Q, \mathcal H_t)$, which quantifies the remaining uncertainty after round $t$. Each retrieval step reduces entropy by:
\begin{equation}
V_{t-1} - V_t = I(A^\star; \mathcal P_t \mid Q, \mathcal H_{t-1}),
\end{equation}
We focus on the first step of iterative RAG that $t=1$. The entropy reduction for the first step is 
\begin{equation}
    V_{0} - V_1 = I(A^\star; \mathcal P_1 \mid Q, \mathcal H_{0}).
\end{equation}
In Model-Initialized thinking, $\mathcal{H}_0=\{\emptyset\}$ while in our Early Knowledge Alignment, the $\mathcal{H}_0=\{\mathcal P_{0}\}$. Thus 

Summing over all rounds, the total information gain of the adaptive strategy satisfies:
\begin{align}
\mathbb{E}_\pi \left[ I(A^\star; \mathcal H_T^{EKA} \mid Q) \right] &= \mathbb{E}_\pi \left[\sum_{t=1}^T I(A^\star; \mathcal P_t^{EKA} \mid Q, \mathcal H_{t-1}^{EKA})\right] \\
&\ge  \mathbb{E}_\pi \left[\sum_{t=1}^T I(A^\star; \mathcal P_t \mid Q, \mathcal H_{t-1})\right] \\
& = \mathbb{E}_\pi \left[ I(A^\star; \mathcal H_T \mid Q) \right],
\end{align}
while the unequality comes from the fact that with $\mathcal{H}_0=\{\mathcal P_{0}\}$, which is highly related to $Q$, at each step $t>=1$,
\begin{equation}
    I(A^\star; \mathcal P_t^{EKA} \mid Q, \mathcal H_{t-1}^{EKA}) \ge   I(A^\star; \mathcal P_t \mid Q, \mathcal H_{t-1}), \\
\end{equation}
which means the EKA is no worse than the traditional thinking. 

Let $\rho_T$ denote the information gain per token at the end of the iterative operation:
\begin{equation}
\rho_T = \frac{I(A^\star; \mathcal H_T \mid Q)}{B},
\end{equation}
From a Bayesian viewpoint, retrieval efficiency can be seen as how much uncertainty is reduced per token. EKA achieves a greater entropy reduction under the same budget, or requires fewer tokens to reach the same posterior certainty, it is strictly more efficient.
Moreover, by Fano’s inequality,
\begin{equation}
P_e \le \frac{H(A^\star \mid Q) - I(A^\star; \mathcal H_T \mid Q) + 1}{\log |\mathcal A|},
\end{equation}
we conclude that the lower the conditional entropy, the lower the expected error. Therefore, greater mutual information directly translates into improved answer accuracy.

In conclusion, Early Knowledge Alignment enables the agent to get more information gain and lower entropy at the end of iterative RAG, leading to more efficient and accurate question answering.
\end{proof}

\section{Detailed Implementations and Hyperparameters}
\label{appendix:detailedimplementation}

\subsection{Baselines in Graph-R1 Setting}
Baselines in Graph-R1 setting first utilizes \textbf{GPT-4o-mini} as the inference-only generator. This includes \textbf{NaiveGeneration}, which performs zero-shot generation without retrieval to evaluate the base model's capacity, and \textbf{StandardRAG} \citep{RAG}, a conventional chunk-based retrieval-augmented generation approach. We also include several graph-based retrieval methods: \textbf{GraphRAG} \citep{GraphRAG}, which constructs entity graphs for one-shot retrieval; \textbf{LightRAG} \citep{LightRAG}, a lightweight variant that builds compact graphs for more efficient retrieval; \textbf{PathRAG} \citep{PathRAG}, which performs retrieval via path-based pruning on entity graphs; \textbf{HippoRAG2} \citep{HippoRAG2}, which employs a hierarchical path planner over knowledge graphs to improve retrieval efficiency; and \textbf{HyperGraphRAG} \citep{HyperGraphRAG}, which constructs n-ary relational hypergraphs to support a single retrieval step.

The second set of baselines is based on the \textbf{\texttt{Qwen2.5-Instruct} (7B)} model. We begin with foundational methods, including a \textbf{NaiveGeneration} approach as a lower-bound, the classic \textbf{StandardRAG} \citep{RAG} pipeline, and \textbf{SFT} \citep{SFT}, which involves supervised fine-tuning on QA pairs. Furthermore, we evaluate several advanced methods trained with reinforcement learning: \textbf{R1} \citep{GRPO}, a GRPO-trained policy that generates answers directly without retrieval; \textbf{Search-R1} \citep{Search-R1}, a multi-turn chunk-based retrieval method trained with GRPO; \textbf{R1-Searcher} \citep{R1-Searcher}, a two-stage GRPO-based method for chunk-based retrieval; and \textbf{Graph-R1}\citep{Graph-R1}, an agentic GraphRAG framework enhanced by end-to-end reinforcement learning.

\vspace{-1.5mm}
\subsection{Baselines In Search-R1 Setting}
In Search-R1 setting, despite the baselines in last section, we also compare against prominent reasoning and generation strategies: \textbf{CoT} \citep{wei2022chain}: reasoning with chain of thought; \textbf{IRCoT}\citep{trivedi2022interleaving}: reasoning with chain of thought with retreival; \textbf{Search-o1}\citep{li2025search}: integrating an agentic search workflow into the reasoning process; and \textbf{Rejection Sampling}\citep{ahn2024large}: SFT on trajectories that succeed.
\subsection{Metrics}


\noindent\textbf{Exact Match (EM).}
This metric provides a strict evaluation of answer accuracy. It determines if the generated answer $y_i$ is identical to the ground-truth reference $y_i^\star$ after both have undergone a normalization process. This process typically includes lowercasing, removing punctuation, and standardizing whitespace. The score is 1 if they match perfectly, and 0 otherwise. The final EM score is the average over all samples:
\begin{equation}
  \text{EM} = \frac{1}{N} \sum_{i=1}^{N} \mathbb{I} \left\{ \text{norm}(y_i) = \text{norm}(y_i^\star) \right\}.
\end{equation}

\noindent\textbf{F1 Score.} 
Unlike the all-or-nothing EM, the F1 score offers a more nuanced measure of quality by assessing the word-level (token) overlap between the prediction and the ground truth. It calculates the harmonic mean of precision (the fraction of predicted tokens that are correct) and recall (the fraction of ground-truth tokens that are predicted), providing a balanced assessment of token accuracy:
\begin{equation}
  \text{F1} = \frac{1}{N} \sum_{i=1}^{N} \frac{2 \cdot |\text{tokens}(y_i) \cap \text{tokens}(y_i^\star)|}{|\text{tokens}(y_i)| + |\text{tokens}(y_i^\star)|}.
\end{equation}

\noindent\textbf{Retrieval Similarity (R-S).}
This metric evaluates the quality of the retrieval component of the RAG system, rather than the final generated answer. It measures the semantic relevance of the retrieved context $k_{\text{retr}}^{(i)}$ compared to the ideal "gold" context $k_{\text{gold}}^{(i)}$. To do this, both texts are converted into vector representations using a semantic embedding function $\text{Enc}(\cdot)$, and their cosine similarity is computed:
\begin{equation}
  \text{R-S} = \frac{1}{N} \sum_{i=1}^{N} \cos\left(\text{Enc}(k_{\text{retr}}^{(i)}), \text{Enc}(k_{\text{gold}}^{(i)})\right).
\end{equation}

\subsection{Hyperparameters}
We show in Table \ref{tab:hyperparams} the hyperparameters in Graph-R1 setting. In Search-R1 setting, the hyperparameters are the same as Search-R1. The models with EKA share the same hyperparameters with the backbone method.

\begin{table}[h]
\centering
\fontsize{7pt}{7.5pt}\selectfont
\setlength{\tabcolsep}{2.8mm}{
\begin{tabular}{lcccccc}
\toprule
\textbf{Method} & \textbf{Backbone} & \textbf{Batch Size} & \textbf{Max Length} & \textbf{Top-K} & \textbf{Algo} & \textbf{Epochs} \\
\midrule
NaiveGeneration     & Qwen2.5 / GPT-4o-mini & --   & $\infty$ & N/A  & --     & --  \\
StandardRAG         & Qwen2.5 / GPT-4o-mini & --   & $\infty$ & 5 Chunks  & --     & --  \\
GraphRAG            & GPT-4o-mini           & --   & $\infty$ & 60 & --     & --  \\
LightRAG            & GPT-4o-mini           & --   & $\infty$ & 60  & --     & --  \\
PathRAG             & GPT-4o-mini           & --   & $\infty$ & 60  & --     & --  \\
HippoRAG2           & GPT-4o-mini           & --   & $\infty$ & 60  & --     & --  \\
HyperGraphRAG       & GPT-4o-mini           & --   & $\infty$ & 60  &  --     & --  \\
SFT                 & Qwen2.5 (7B)              & 16   & 4096 & N/A  & LoRA     & 3   \\
\rowcolor{R1!15} R1                  & Qwen2.5 (7B)               & 128   & 4096 & N/A  & GRPO   & 3   \\
\rowcolor{Search-R1!15} Search-R1           & Qwen2.5 (7B)& 128   & 4096 & 5 Chunks / Turn & GRPO & 6\\
 \rowcolor{Search-R1!15} Search-R1-PPO& Qwen2.5 (7B)& 128   & 4096 & 5 Chunks / Turn & PPO&10\\
\rowcolor{R1-Searcher!15} R1-Searcher         & Qwen2.5 (7B)               & 128   & 4096 & 5 Chunks / Turn & GRPO & 3 \\
 \rowcolor{R1-Searcher!15} Graph-R1& Qwen2.5 (7B)               & 128   & 4096 & 5 Chunks / Turn & GRPO &3 \\
\bottomrule
\end{tabular}}
\vspace{-1mm}
\caption{\label{tab:hyperparams}
Hyperparameter settings in Graph-R1 setting.}
\vspace{-2mm}
\end{table}

\end{document}